\documentclass{article}

\usepackage{arxiv}

\usepackage[utf8]{inputenc} 
\usepackage[T1]{fontenc}    
\usepackage{hyperref}       
\usepackage{url}            
\usepackage{booktabs}       
\usepackage{amsmath}        
\usepackage{amsfonts}       
\usepackage{nicefrac}       
\usepackage{microtype}      
\usepackage{lipsum}         
\usepackage{graphicx}
\usepackage{subcaption}
\usepackage{multirow}
\usepackage{capt-of}
\usepackage{rotating}   
\usepackage{multirow}   
\usepackage{booktabs}   
\usepackage{amssymb}
\usepackage{pdflscape}
\usepackage{multirow}
\usepackage{graphicx}
\usepackage{natbib}
\usepackage{doi}

\title{Pixel-Precise Explainable Stress Indexing: A Semantic Segmentation Framework for Disease Severity Quantification in Field Crops}

\author{
Raunak Kumar \\
Centre of Studies in Resources Engineering\\
Indian Institute of Technology Bombay\\
Mumbai, Maharashtra, India \\
\texttt{raunakraj2003@gmail.com}
\And
Soumyashree Kar\thanks{Corresponding author.} \\
Centre of Studies in Resources Engineering\\
Indian Institute of Technology Bombay\\
Mumbai, Maharashtra, India \\
\texttt{soumyakar@iitb.ac.in}
}

\renewcommand{\shorttitle}{Semantic Segmentation Framework for Disease Severity Quantification in Field Crops}

\hypersetup{
pdftitle={Deep Learning Architectures for Plant Disease Segmentation: A Comparative Study on Apple Leaf and Rice Datasets},
pdfsubject={cs.CV, cs.LG},
pdfauthor={Author One, Author Two},
pdfkeywords={Semantic segmentation, plant disease detection, U-Net, MobileNetV2, SegFormer, deep learning},
}

\begin{document}
\maketitle

\begin{abstract}
Plant diseases, resulting from both biotic and abiotic stresses cause an estimated 20--40\,\% loss in global agricultural yield annually, resulting in economic damages exceeding USD~220 billion. Accurate and scalable stress quantification is essential for precision agriculture, yet traditional manual assessments are labour-intensive and subjective. This paper proposes a unified deep learning pipeline integrating semantic segmentation, regression-based severity estimation, and disease classification. Stress severity is categorised into four levels (Low to Very High) based on the proportion of infected leaf area. Experiments on the Apple Tree Leaf Disease Segmentation dataset (1,641 samples, six classes) evaluate four models: U-Net (MobileNetV2), SegFormer, FCN, and PSPNet. U-Net with MobileNetV2 achieves the best performance with 98.20\,\% pixel accuracy, 0.70 mIoU, and 99.41\,\% detection accuracy at 14.7\,ms per image, making it suitable for real-time use. SegFormer performs competitively (mIoU~0.66), while FCN and PSPNet show lower spatial accuracy (${\sim}0.49$ mIoU). The computed severity index strongly correlates with expert annotations ($r = 0.968$, $R^2 = 0.937$), demonstrating the system's reliability for automated crop monitoring and decision support.
\end{abstract}


\section{Introduction}
\label{sec:introduction}
Agriculture is the foundation of food security and economic stability for much of the world. In India, agriculture and allied sectors contribute approximately 18.3\,\% to the national Gross Domestic Product and directly employ more than 50\,\% of the workforce \citep{mospi2023}. Despite its central role, the agricultural sector faces a persistent and escalating threat from plant diseases and environmental stresses. The Food and Agriculture Organisation of the United Nations estimates that biotic and abiotic plant stresses cause between 20\,\% and 40\,\% losses in global agricultural yield each year, translating to economic damages exceeding United States Dollars 220 billion annually \citep{fao2023}. These losses are projected to intensify in the coming decades as climate change alters precipitation patterns, raises temperatures, and shifts pest and pathogen distributions into previously unaffected growing regions \citep{bebber2013range}. The intersection of a growing global population demanding more food and a shrinking base of productive agricultural land makes the development of intelligent, automated plant health monitoring technologies not merely desirable but critically necessary. Despite considerable advances in image-based plant health monitoring, existing computational pipelines predominantly perform coarse image-level classification and fail to produce the continuous, spatially explicit severity estimates that agronomic decision-making requires. \\Plant stresses are broadly classified into biotic stresses, arising from living organisms such as fungal pathogens, bacterial diseases, viral infections, and insect pests, and abiotic stresses, which encompass non-living environmental perturbations including temperature extremes, drought, waterlogging, salinity, and nutrient deficiencies \citep{atkinson2012interaction}. Both categories manifest visibly on plant tissue through characteristic changes in colour, texture, shape, and area, rendering image-based detection feasible. Understanding why these symptoms appear is equally important. Photosynthesis is exquisitely sensitive to environmental conditions, with peak efficiency near 25$^\circ$C and sharp decline beyond this range \citep{fay1996}. Vapour pressure deficit (VPD), which quantifies the atmospheric gradient driving transpirational water loss, is among the most consequential variables governing plant physiology: as VPD increases, stomatal conductance decreases, restricting CO$_2$ influx and suppressing carbon assimilation \citep{kitashova2023}. \citet{kar2023} demonstrated using high-throughput phenotyping data that chickpea varieties with superior transpiration restriction under high-VPD conditions exhibit stronger drought resilience, confirming that VPD-related physiological traits are heritable and predictable using machine learning models. These disruptions ultimately express themselves as chlorosis, necrosis, wilting, and lesion formation visual signatures that deep learning models can identify and quantify with high precision \citep{mohanty2016using}, though the degree of visual distinctiveness varies considerably across disease types.\\ The conventional approach to disease severity assessment relies on visual inspection by trained plant pathologists who estimate the proportion of affected leaf area. This process suffers from several fundamental limitations: it is labour-intensive, impractical at scale, and subject to inter-observer variability routinely exceeding 20\,\% \citep{aakanksha2022}. Early computational alternatives based on digital image processing including global thresholding, edge-based segmentation, region growing, and watershed transforms produced promising results under controlled laboratory conditions but degraded severely in genuine field images where non-uniform illumination, overlapping leaves, and soil clutter rendered hand-engineered feature extraction unreliable \citep{pethybridge2015}. Statistical machine learning methods such as K-Means clustering, Fuzzy C-Means, and Support Vector Machines improved detection rates by learning decision boundaries from labelled data \citep{jaware2012}. \citet{sibiya2019} integrated colour threshold segmentation with fuzzy logic to automate severity estimation for maize leaf diseases, yet these approaches remained sensitive to initialisation parameters and lacked robustness across diverse environmental conditions and crop species. \citet{zhang2018deep} provided an early systematic review showing that handcrafted features consistently underperformed learned representations on field-collected images across multiple crop species.

\subsection{Deep Learning Approaches for Plant Disease Analysis}
\label{subsec:dl-approaches}
The advent of Convolutional Neural Networks (CNNs) fundamentally transformed plant disease detection by enabling models to learn hierarchical feature representations directly from raw pixel data \citep{lecun2015deep}. \citet{mafukidze2022} demonstrated that architectures including DenseNet, InceptionV3, VGG16, and MobileNet, trained on the PlantVillage dataset, could achieve up to 99\,\% classification accuracy through transfer learning. \citet{liang2019} proposed a multi-task framework on a ResNet-50 backbone achieving 0.99 accuracy for species recognition, 0.98 for disease classification, and 0.91 for severity estimation simultaneously. At the yield prediction level, \citet{khaki2019} showed that a 21-layer Deep Neural Network trained on genotype, weather, and soil data outperformed LASSO, shallow networks, and regression trees, revealing that environmental factors exert considerably stronger influence on yield than genotype alone. \citet{lu2024} subsequently demonstrated that a CNN-LSTM-Attention model integrating satellite-derived vegetation indices and environmental variables achieved $R^2 = 0.80$ for rice, maize, and soybean yield prediction, substantially outperforming standalone Random Forest and XGBoost models. Despite these advances, classification-based approaches assign a single disease label to the entire image, providing no spatial information about affected regions or severity a fundamental limitation that motivated the shift toward semantic segmentation \citep{thenmozhi2019crop}.\\ Semantic segmentation assigns a class label to every individual pixel, enabling precise localisation and area-based quantification of disease extent. \citet{fcn2015} pioneered end-to-end pixel-wise dense prediction, while \citet{unet2015} introduced the encoder-decoder structure with skip connections that remains highly effective for agricultural datasets of limited size. The DeepLab family \citep{deeplab2018} employed atrous convolutions to increase receptive field without resolution loss; \citet{ji2022} combined DeepLabV3+ with fuzzy logic for grape disease severity classification, achieving mean IoU exceeding 85\,\% and overall severity accuracy of 97.75\,\%. The introduction of Vision Transformers \citep{dosovitskiy2020vit} demonstrated that self-attention mechanisms could model long-range spatial dependencies more effectively than convolutional inductive biases. \citet{segformer2021} adapted this for segmentation through a hierarchical transformer encoder and all-MLP decoder, consistently outperforming CNN-based models on agricultural benchmarks \citep{zhou2025}. \citet{wang2021knowledge} further demonstrated that knowledge-guided deep learning architectures could incorporate domain-specific plant pathology priors to improve segmentation accuracy under low-data regimes.

\subsection{Data-Efficient and Multimodal Approaches}
\label{subsec:data-efficient}
A critical bottleneck across all these paradigms remains the acquisition of large quantities of pixel-wise annotated field images, which is prohibitively expensive and time-consuming. Self-supervised learning approaches such as the Masked Autoencoder \citep{he2022mae} and DINO \citep{caron2021dino} address this by learning transferable representations from unlabelled data. \citet{kar2023_ssl} further demonstrated that self-supervised pretraining (e.g., NNCLR, BYOL, and Barlow Twins) enables robust and annotation-efficient classification of agricultural insect pests, outperforming conventional transfer learning approaches, particularly under low-data regimes. \citet{bedi2024} achieved 94.54\,\% mIoU with only two training samples per class, demonstrating the potential of data-efficient learning in agricultural applications. Sensor fusion approaches integrating thermal, hyperspectral, and environmental sensor data with RGB imagery have further demonstrated consistent gains over single-modality baselines \citep{kamarudin2023,gold2020spectral}, pointing toward next-generation precision agriculture platforms that fuse visual symptom data with physiological sensor readings and satellite-derived vegetation indices to generate holistic crop health assessments \citep{west2003}.\\ The distinction between disease detection and disease quantification carries direct agronomic significance. A binary classification output diseased vs.\ healthy provides no information about the severity of infection or the urgency of intervention. In contrast, a continuous Stress Severity Index (SSI) expressed as the proportion of visibly diseased leaf tissue enables threshold-based management decisions: international plant health standards such as the EPPO rating scale and USDA disease severity guidelines use precisely such proportional metrics to calibrate spray timing, application rate, and economic threshold calculations \citep{eppo2021,horsfall1942}. Automating SSI estimation through semantic segmentation thus addresses a concrete gap between current deep learning capability and field-deployable precision-agriculture tooling.

\subsection{Research Gaps and Contributions}
\label{subsec:research-gaps}
Despite significant progress in deep learning for plant disease detection, several critical gaps limit practical deployment. The most pervasive is the overemphasis on image-level classification, which identifies disease type but provides no spatial distribution or fractional area information both essential for computing agronomically meaningful severity indices. Most published datasets, including PlantVillage \citep{hughes2015open}, consist of controlled laboratory images with uniform backgrounds, creating a domain gap that causes severe performance degradation under real field conditions. Where spatial analysis is attempted, bounding-box detection inflates severity estimates by including healthy tissue, whereas true quantification demands pixel-level semantic segmentation. The scarcity of pixel-wise annotated field datasets further limits supervised segmentation training, and existing works treat segmentation, severity regression, and disease classification as separate tasks requiring distinct models \citep{gonccalves2021}. Finally, state-of-the-art architectures such as DeepLabV3+ and Mask2Former, with tens of millions of parameters, remain impractical for UAV and low-cost edge deployment \citep{chlingaryan2018machine}.\\ This work directly addresses these gaps through four core contributions. (I) We develop an end-to-end pipeline that simultaneously performs pixel-wise disease segmentation, computes a continuous Stress Severity Index (SSI) as the ratio of diseased to total leaf pixels, and identifies disease type within a single trainable architecture. (II) We introduce a hybrid U-Net with MobileNetV2 encoder that achieves state-of-the-art accuracy on the ATLDS and RiceSEG dataset at significantly reduced parameter count and inference time, enabling edge and UAV deployment. (III) We conduct a systematic comparison of four architectures U-Net with MobileNetV2, SegFormer, FCN, and PSPNet providing actionable guidance on the accuracy-efficiency trade-off for field deployment. Fourth, SSI predictions are validated against expert annotations, yielding a Pearson correlation of $r = 0.968$ and $R^2 = 0.937$, confirming the framework as a reliable proxy for manual expert assessment.

\section{Data Description}
\label{sec:data-description}

\subsection{Apple Tree Leaf Disease Segmentation (ATLDS) Dataset}
\label{subsec:atlds}
\begin{figure}[!t]
    \centering
    \includegraphics[width=\linewidth]{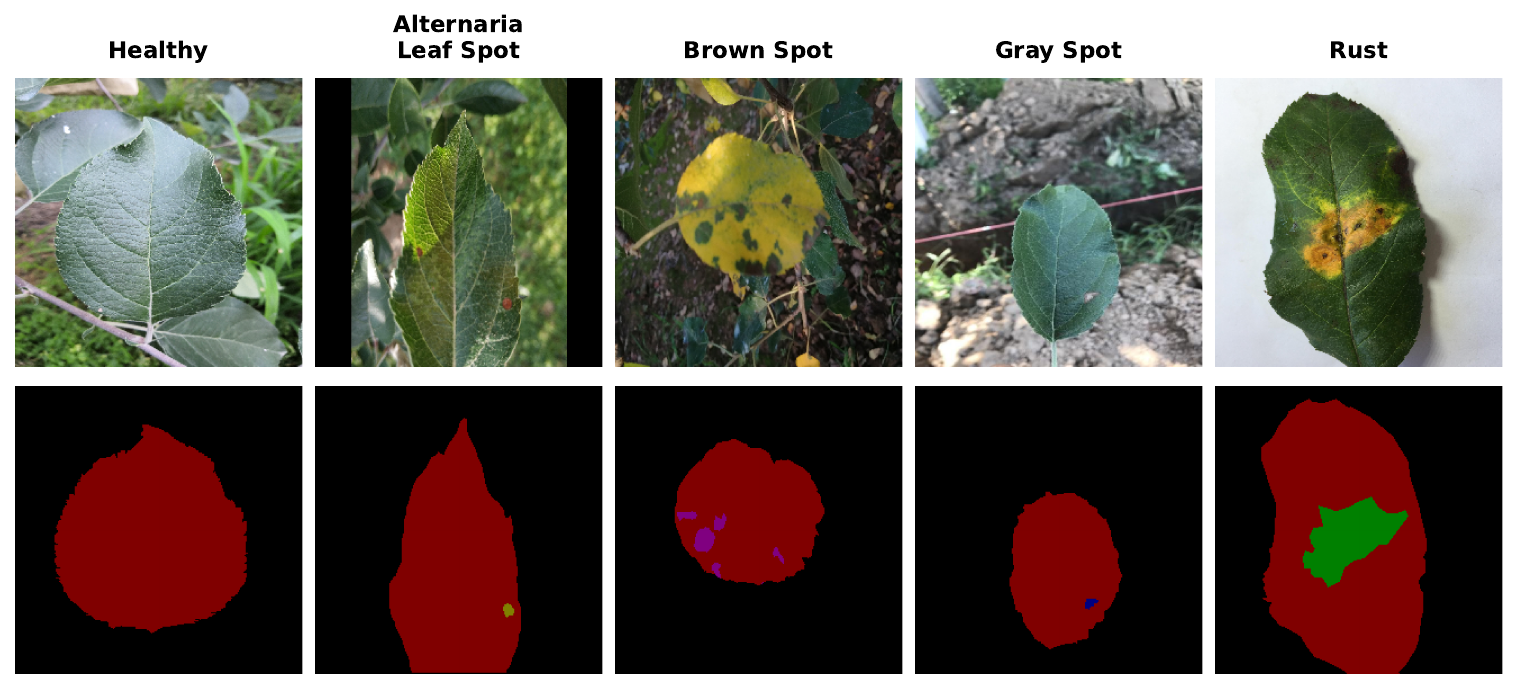}
    \caption{Representative samples from the ATLDS dataset showing
    RGB images and ground-truth segmentation masks.}
    \label{fig:sample_grid_apple}
\end{figure}

\begin{figure}[!t]
    \centering
    \includegraphics[width=0.48\linewidth]{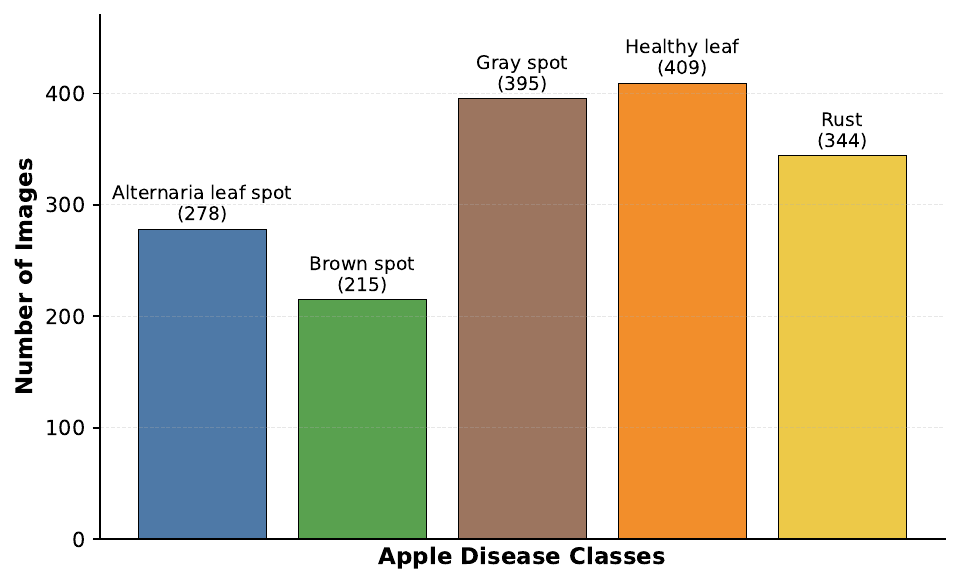}
    \hfill
    \includegraphics[width=0.48\linewidth]{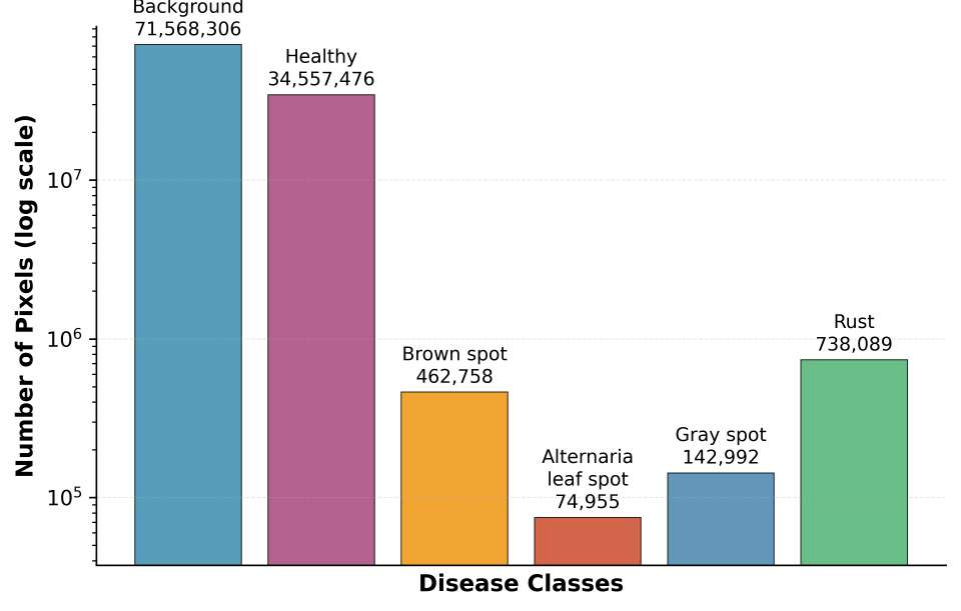}
    \caption{Class distribution of the ATLDS dataset: number of images
    per class (left) and pixel-intensity distribution per class on a
    logarithmic scale (right). Background and Healthy pixels dominate;
    Brown Spot and Alternaria Leaf Spot are severe minority classes.}
    \label{fig:class_dist}
\end{figure}

The primary dataset used for all model training and evaluation is the Apple Tree Leaf Disease Segmentation (ATLDS) datasetFig.\ref{fig:sample_grid_apple} \citep{feng2021apple}. It comprises 1,641 paired RGB image mask samples, where each image is accompanied by a pixel-level segmentation mask annotated by plant pathology experts using the LabelMe annotation platform. The images were collected under natural field conditions, encompassing a wide range of lighting conditions, leaf orientations, background complexity, and disease progression stages. The dataset spans six semantic classes: Background, Healthy Leaf (409 samples), Gray Spot (395 samples), Rust (344 samples), Alternaria Leaf Spot (278 samples), and Brown Spot (215 samples). The dataset is a mixture of field-acquired images with natural backgrounds and close-up laboratory-style images, a deliberate diversity that increases intra-class variability and substantially improves the generalisation of trained models to unseen field conditions \citep{barbedo2018factors}.\\ As illustrated in Fig.\,\ref{fig:class_dist}, the dataset exhibits pronounced class imbalance: the Healthy Leaf class contains nearly twice as many samples as Brown Spot, and at the pixel level, Background and Healthy pixels dominate the total pixel count by several orders of magnitude, while disease-affected pixels form a sparse minority. This imbalance is a critical training challenge addressed through class weighting and focal loss strategies described in Section\,\ref{subsubsec:loss-ablation}.


\subsection{Global Rice Multiclass Segmentation Dataset (RiceSEG)}
\label{subsec:riceseg}
\begin{figure}[!t]
    \centering
    \includegraphics[width=\linewidth,height=10cm]{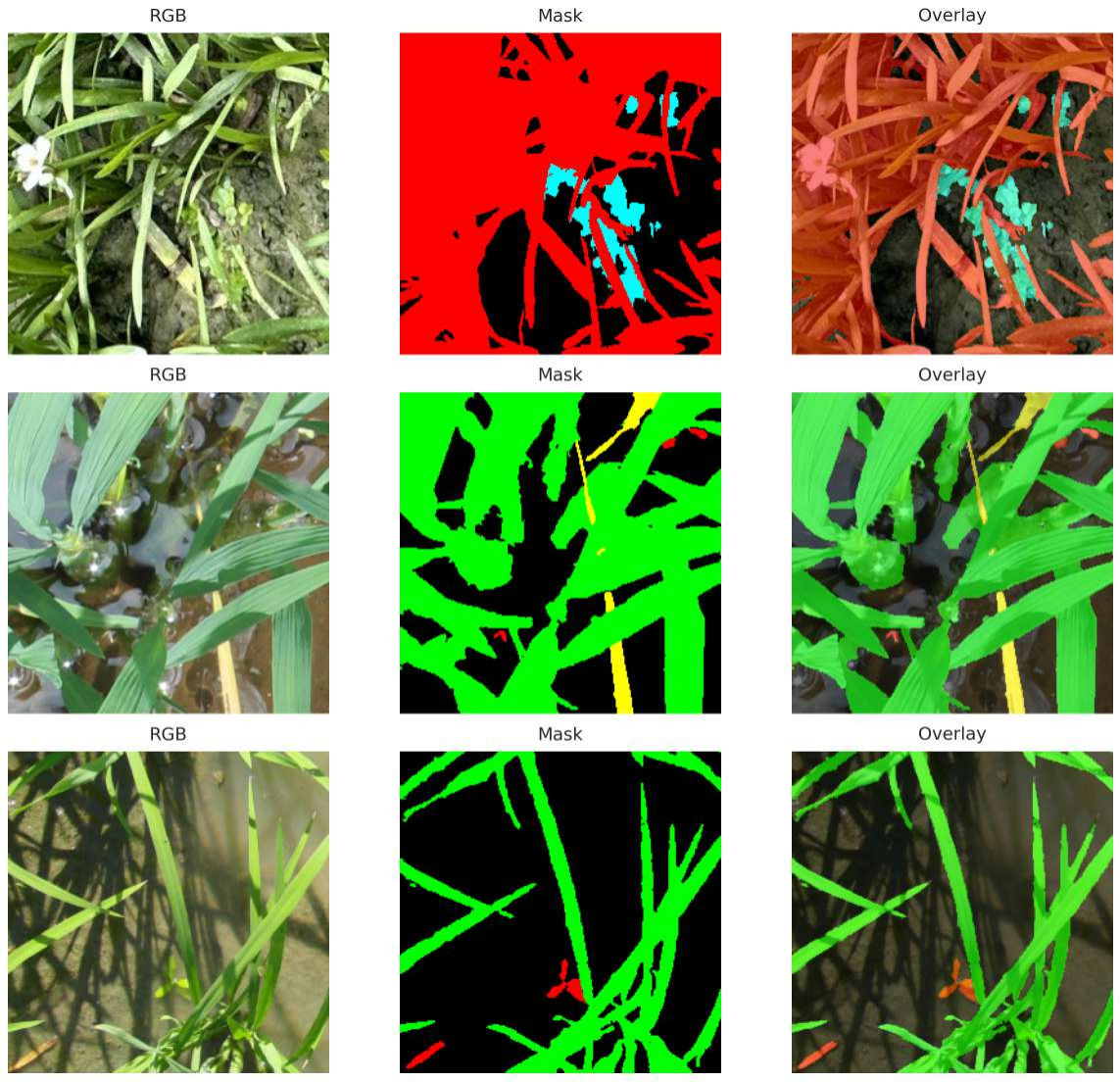}
    \caption{Representative samples from the RiceSEG dataset showing
    RGB images, ground-truth segmentation masks, and overlay
    visualisations.}
    \label{fig:sample_grid}
\end{figure}

\begin{figure}[!t]
    \centering
    \includegraphics[width=\linewidth,height=8cm]{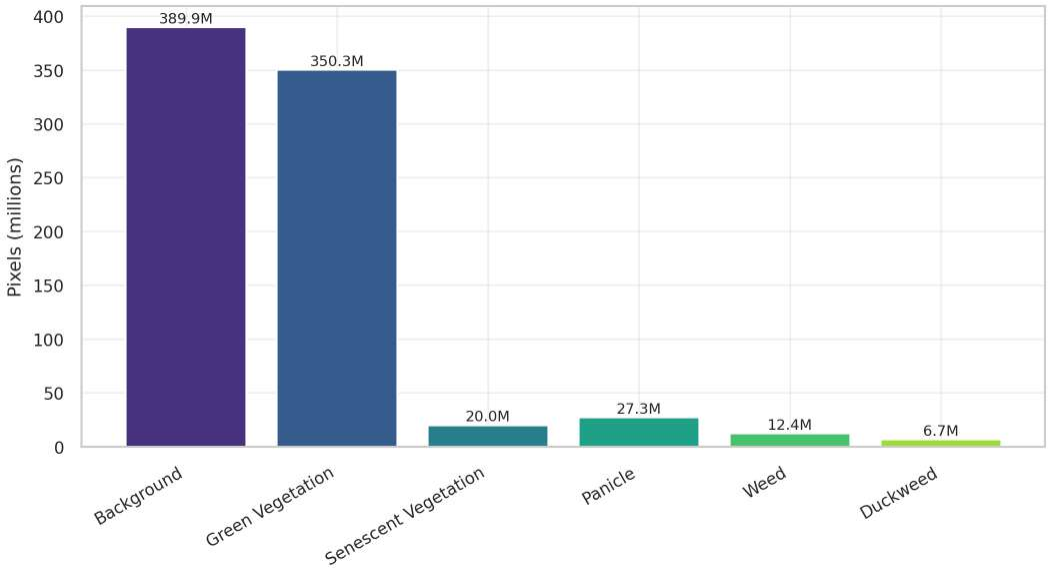}
    \caption{Pixel-wise class distribution of the RiceSEG dataset
    highlighting class imbalance across different categories.}
    \label{fig:pixel_distribution}
\end{figure}

The inclusion of the RiceSEG dataset serves a specific and deliberate methodological purpose: to evaluate the generalisability of the proposed architecture beyond single-leaf close-up imagery to wide-field canopy segmentation under substantially different imaging conditions. While ATLDS consists predominantly of isolated leaf images against simple backgrounds, RiceSEG contains complex, multi-species field scenes with overlapping canopies, water surfaces, and spatially heterogeneous vegetation at the plot scale. This structural contrast allows us to assess whether the skip-connection spatial recovery that drives the proposed model's lesion boundary precision on ATLDS translates to robust canopy-level segmentation where class boundaries are diffuse rather than sharp. It is important to note that the SSI framework described in Section~\ref{sec:methodology} is evaluated exclusively on the ATLDS dataset, as RiceSEG does not provide pixel-level disease severity ground-truth annotations; the RiceSEG evaluation instead serves as a domain-shift robustness benchmark.\\ The RiceSEG dataset \citep{zhou2025} is a large-scale, publicly available benchmark designed for multiclass semantic segmentation of rice crops under real-world field conditions. It consists of pixel-level annotated images capturing multiple classes such as background, green vegetation, senescent vegetation, panicle, weeds, and duckweed. The dataset spans diverse environmental conditions, growth stages, and imaging setups, making it representative of practical agricultural scenarios \citep{lu2022transformer}.\\ The dataset was collected from multiple geographic regions across Asia and Africa, including countries such as China, Japan, India, the Philippines, and Tanzania, using a variety of imaging devices such as DSLR cameras and smartphones. Images were captured under varying viewpoints and illumination conditions, introducing natural variability in scale, occlusion, and background complexity. This diversity makes the dataset particularly suitable for evaluating model robustness in unconstrained environments. Figure~\ref{fig:sample_grid} presents representative samples from the dataset, including RGB images, corresponding ground-truth masks, and overlay visualisations. The examples highlight challenges such as dense canopy structures, overlapping leaves, presence of water, and background noise, all of which complicate accurate pixel-wise classification. To better understand the dataset characteristics, an exploratory data analysis (EDA) was performed. Figure~\ref{fig:pixel_distribution} shows the pixel-wise class distribution across the dataset. It can be observed that the dataset is highly imbalanced, with background and green vegetation dominating the majority of pixels, while classes such as weeds and duckweed occupy significantly smaller proportions.

\section{Methodology}
\label{sec:methodology}

\subsection{Overview}
\label{subsec:overview}
\begin{figure*}[!htbp]
    \centering
    \includegraphics[width=\textwidth]{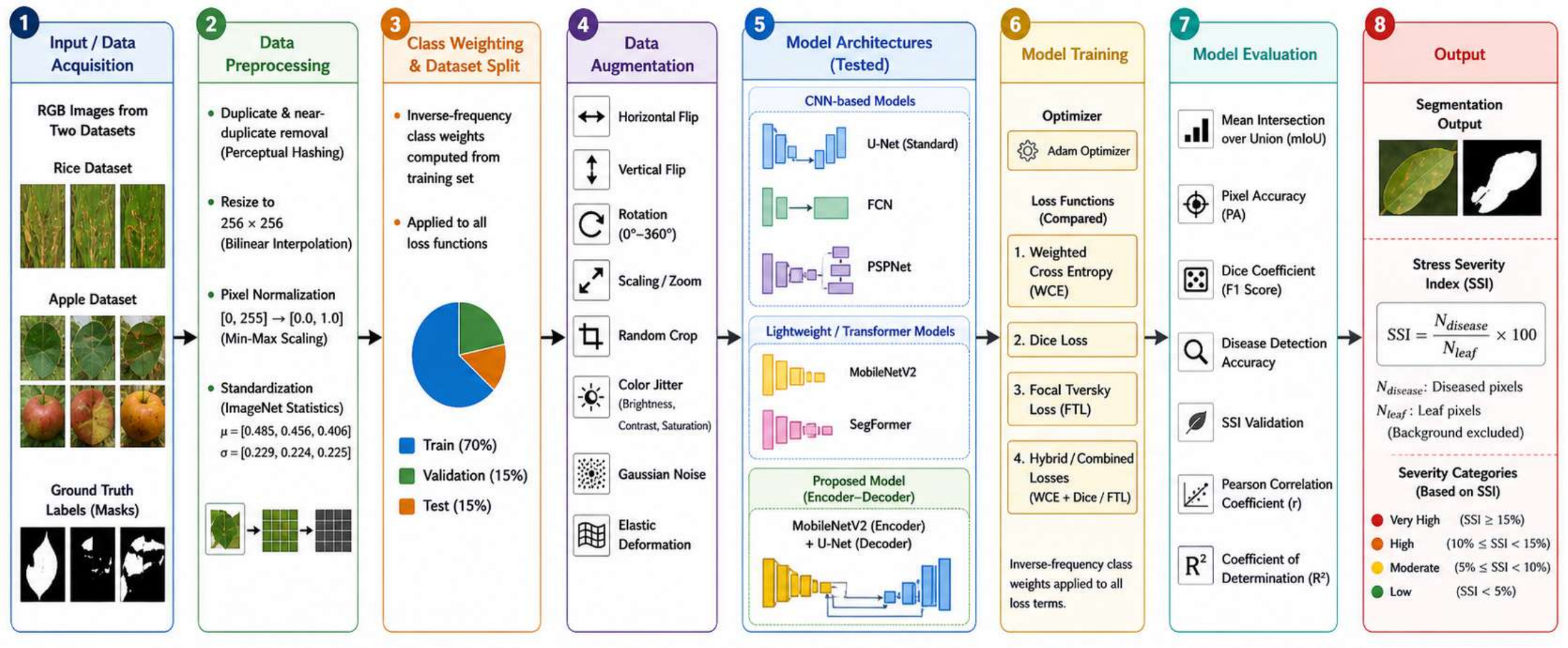}
    \caption{End-to-end workflow of the proposed framework. The Data
    Preparation stage includes data collection, expert annotation,
    preprocessing, augmentation, and dataset splitting. Model
    Development involves training and optimisation, followed by
    semantic segmentation and regression/classification tasks. The
    final output consists of a segmentation mask, a computed severity
    (SSI) value, and a disease type label.}
    \label{fig:workflow}
\end{figure*}
The operational workflow of the proposed system, illustrated in Fig.\,\ref{fig:workflow}, proceeds through eight sequential stages. In the first stage, field images from public repositories (as described in the earlier section) are curated. These images were captured using UAV platforms and ground-level digital cameras under healthy, mild-stress, and severe-stress conditions, covering disease manifestations, nutrient deficiencies, and water shortage scenarios across multiple growth stages. The second stage involves annotation or labeling of pixel-wise stress regions using LabelMe \citep{torralba2010labelme}; all annotations were independently cross-validated to ensure label consistency and accuracy. The third stage applies the preprocessing pipeline described in Section\,\ref{subsec:preprocessing}, including quality filtering, resizing to $256\times256$, and per-channel standardisation. The fourth stage performs online data augmentation during training. The fifth stage trains the selected segmentation architecture using the combined loss of Eq.\,(\ref{eq:totalloss}) with Adam optimiser, initial learning rate $10^{-4}$, batch size 8, up to 50 epochs, Early Stopping with patience 15, and ReduceLROnPlateau scheduler with reduction factor 0.5 and patience 8. The sixth stage applies the trained model to generate pixel-wise segmentation masks for test images. The seventh stage computes the Stress Severity Index using Eq.\,(\ref{eq:ssi}) and assigns the disease type using Eq.\,(\ref{eq:diseasetype}). The eighth and final stage evaluates all metrics on the held-out test partition.

\subsection{Preprocessing Workflow}
\label{subsec:preprocessing}
A systematic multi-stage preprocessing pipeline was applied to all images prior to training \citep{shorten2019survey}. Images exhibiting more than 50\,\% background coverage, motion blur, overexposure, or significant sensor noise were manually identified and excluded. Duplicate and near-duplicate images were detected using perceptual hashing and removed to prevent data leakage between training and test splits. All retained images were resized to a uniform resolution of $256 \times 256$ pixels using bilinear interpolation. Pixel intensities were then normalised from the native range $[0, 255]$ to $[0.0, 1.0]$ using per-channel min-max scaling, followed by standardisation using the ImageNet channel-wise statistics (mean $\mu = [0.485, 0.456, 0.406]$ and standard deviation $\sigma = [0.229, 0.224, 0.225]$) to align the input distribution with the pretrained MobileNetV2 and SegFormer encoders. The same spatial transformations applied to each image were identically applied to its corresponding segmentation mask to preserve pixel-label correspondence throughout.

\subsection{Augmentation Strategies}
\label{subsec:augmentation}
Data augmentation was applied online during training using a stochastic pipeline to artificially increase the effective dataset size and expose the model to a wider range of realistic imaging variations, thereby reducing overfitting and improving generalisation to unseen field conditions \citep{shorten2019survey}. Random horizontal and vertical flipping was applied with a probability of 0.5 each, simulating the natural variability in leaf orientation and camera angle. Random rotation within $\pm30^\circ$ was applied to make the model robust to arbitrary leaf tilt. Random scaling and cropping within a scale range of $[0.70, 1.00]$ of the original image, followed by resizing back to $256 \times 256$, simulated the effect of varying camera distances and partial occlusion. Brightness adjustment with a random factor from $[0.7, 1.3]$ modelled the wide variation in natural sunlight intensity. Contrast adjustment with a factor from $[0.8, 1.2]$ and random colour jittering of hue and saturation improved robustness to variations in plant pigmentation. Gaussian noise with standard deviation uniformly sampled from $[0.0, 0.02]$ simulated sensor noise and atmospheric scattering. All augmentations were applied identically to both the RGB image and its corresponding segmentation mask.

\subsection{SSI Threshold Definition}
\label{subsec:ssi-threshold}
The severity threshold boundaries of 5\,\%, 10\,\%, and 15\,\% diseased leaf area are grounded in established plant pathology practice. The Horsfall Barratt rating scale \citep{horsfall1942}, the most widely adopted ordinal severity reference in applied plant pathology, uses a logarithmic gradation in which the transition from visually detectable infection to economically significant infection occurs in the 5--15\,\% affected area range for foliar diseases. Complementarily, the European and Mediterranean Plant Protection Organisation (EPPO) standard PP~1/181 for disease assessment in fruit crops identifies the 5\,\% and 15\,\% thresholds as the boundaries between the \textit{low}, \textit{moderate}, and \textit{high} economic risk categories relevant to spray-timing decisions \citep{eppo2021}. The selected boundaries therefore align with internationally recognised agronomic intervention thresholds rather than being arbitrary computational choices.

\subsection{Comparison of Deep Learning Model Architectures}
\label{subsec:model-comparison}
Semantic segmentation for plant disease quantification requires architectures that balance three competing demands: fine-grained spatial resolution for accurate lesion boundary delineation, global contextual reasoning for distinguishing visually similar disease classes, and computational efficiency for deployment on UAV-mounted or edge-processing hardware \citep{chen2023survey}. The six architectures evaluated in this study represent distinct design philosophies along this continuum, ranging from classical fully convolutional pipelines to lightweight hybrid encoder decoder networks and transformer-based models.

\subsubsection{FCN (Baseline)}
\label{subsubsec:fcn}
The Fully Convolutional Network \citep{fcn2015} adapts a VGG-16
backbone for dense per-pixel prediction by replacing fully connected
layers with $1\times1$ convolutions, fusing multi-scale predictions
via bilinear upsampling in its FCN-8s variant. The backbone uses
$3\times3$ kernels (occasionally $7\times7$ in the first layer) with
$2\times2$ max-pooling for downsampling; the original formulation
omits batch normalization, though modern reimplementations often
include it. FCN established the
foundational paradigm for end-to-end pixel-wise dense prediction and
serves as the primary baseline in this study.

\subsubsection{PSPNet}
\label{subsubsec:pspnet}
Pyramid Scene Parsing Network \citep{pspnet2017} extends a ResNet-50 backbone with a
Pyramid Pooling Module that aggregates global context at four spatial
scales ($1\times1$, $2\times2$, $3\times3$, and $6\times6$),
capturing scene-level semantics alongside local features. The backbone uses $3\times3$ and dilated
convolutions (rates 2/4) with batch normalization throughout;
$2\times2$ max-pooling is used in the backbone, while the pyramid
pooling module itself uses adaptive average pooling.

\subsubsection{U-Net}
\label{subsubsec:unet}
U-Net \citep{unet2015} introduced the encoder-decoder structure with
skip connections that remains highly effective for agricultural
datasets of limited size. The encoder progressively downsamples the
input through repeated blocks of $3\times3$ convolutions (with
$1\times1$ convolutions at the output layer) and $2\times2$
max-pooling, with batch normalization applied after most conv
blocks, building a hierarchy of feature representations at
increasing semantic abstraction. The decoder symmetrically upsamples
these representations back to the original resolution via transposed
convolution/upsampling, at each level concatenating the corresponding
encoder feature map via skip connections to reintroduce spatial
detail lost during downsampling. This direct reuse of encoder
features allows the decoder to precisely reconstruct object
boundaries and fine lesion edges that would otherwise be irretrievably
lost through successive pooling operations \citep{zhou2018unet++}.

\subsubsection{MobileNetV2}
\label{subsubsec:mobilenetv2}
The MobileNetV2 model \citep{sandler2018mobilenetv2} employs
inverted residual blocks with linear bottlenecks as its fundamental
computational unit. For an input feature tensor
$\mathbf{F} \in \mathbb{R}^{H \times W \times d_{\mathrm{in}}}$:
\begin{equation}
  \mathbf{F'} \;=\;
  \mathrm{PW}_{\downarrow}\!\!\left(
    \mathrm{DW}\!\left(
      \mathrm{PW}_{\uparrow}(\mathbf{F})
    \right)
  \right) + \mathbf{F},
  \label{eq:invres}
\end{equation}
where residual addition of $\mathbf{F}$ to the transformed
features produces the output tensor $\mathbf{F'}$, enabling efficient
gradient propagation while preserving low-level image information. $\mathrm{PW}_\uparrow$ expands channels by factor $t$ using a
$1\times1$ pointwise convolution, $\mathrm{DW}$ applies a $3\times3$
depthwise convolution operating independently on each channel, and
$\mathrm{PW}_\downarrow$ projects back to the original channel
dimension linearly (without a non-linearity). This linear projection
avoids information collapse in low-dimensional manifolds, which is
particularly important for preserving subtle disease-related colour and
texture gradients. Depthwise separable convolutions reduce the
computational cost of a standard convolution by a factor of
$1/k^2 + 1/d_{\mathrm{out}}$ for kernel size $k$, accounting for the
substantial inference efficiency advantage. Feature maps are extracted
at strides $\{4, 8, 16, 32\}$ from successive inverted residual
blocks to form the four skip connection sources
$\mathbf{S}_1$--$\mathbf{S}_4$ used by the U-Net decoder.

\subsubsection{SegFormer}
\label{subsubsec:segformer}
SegFormer \citep{segformer2021} departs from purely convolutional
designs, using a hierarchical Transformer encoder with overlapping
patch embeddings (patch size $\sim7$, stride 4 in the first stage;
$3\times3$, stride 2 in later stages) instead of convolutional
kernels or pooling layers, with downsampling achieved via strided
patch merging. LayerNorm is used throughout, rather than BatchNorm.
The decoder is a lightweight all-MLP head aggregating multi-scale
encoder features without additional convolutions or pooling.

\subsubsection{Proposed Model: U-Net with MobileNetV2 Backbone}
\label{subsubsec:proposed-architecture}
\begin{figure*}[t]
\centering
\includegraphics[height=8cm]{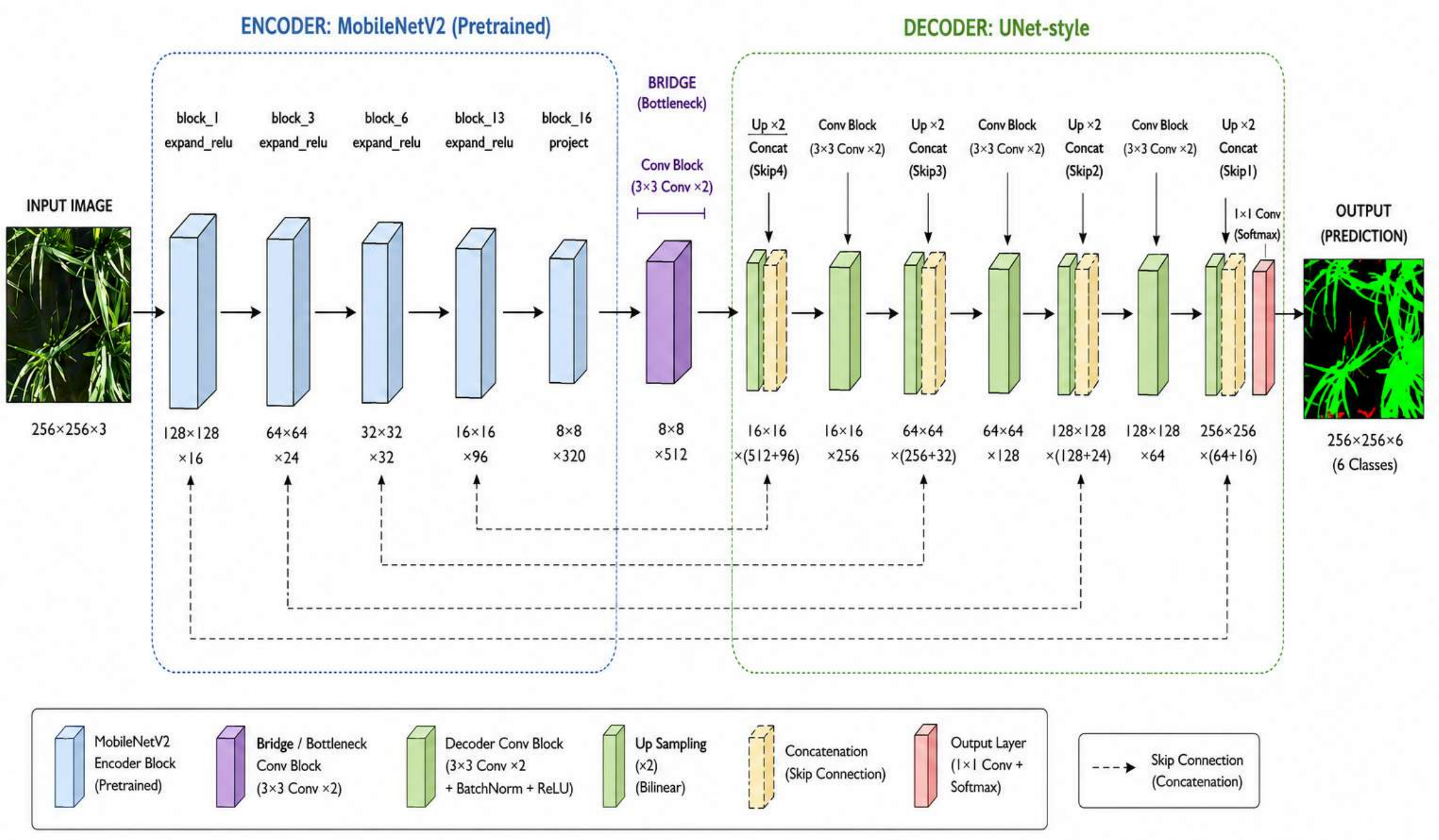}
\caption{U-Net with MobileNetV2 encoder showing encoder, bridge,
and decoder with skip connections, producing a segmentation map.}
\label{fig:unet_arch}
\end{figure*}
The primary proposed architecture integrates the U-Net framework with
a MobileNetV2 encoder pretrained on ImageNet, combining precise
spatial recovery through skip-connection decoding with the efficient,
lightweight depthwise separable convolutional backbone. The key
novelty lies in this combination for agricultural disease
segmentation: a model that is simultaneously accurate, computationally
efficient, and well-suited for limited-data field settings, deployable
on edge and UAV platforms at 14.7\,ms per image with only 9.69\,M
parameters a 68\,\% reduction over standard U-Net.
\paragraph{Intermediate Bridge Module.}
The Bridge module applies two successive $3\times3$ convolutional
blocks to the deepest encoder output (stride 32), producing an
$8\times8\times512$ semantic bottleneck. This module captures global
disease context disease extent, dominant class distribution, and
scene-level colour statistics before spatial recovery begins in
the decoder. The two convolutional blocks include Batch Normalisation
and ReLU activation to stabilise training and prevent feature map
saturation at the semantic bottleneck.

\paragraph{Skip Connections and Decoder Logic.}
The four-level decoder progressively doubles spatial resolution at
each stage via bilinear upsampling, concatenates the corresponding
skip features $\mathbf{S}_k$ from the MobileNetV2 encoder, and
refines the concatenated representation with two $3\times3$
convolutional blocks with Batch Normalisation and ReLU. The skip
connection at stride 4 ($\mathbf{S}_1$) carries fine edge gradients
and colour boundaries critical for precise lesion boundary
localisation. Deeper skips ($\mathbf{S}_2$--$\mathbf{S}_4$) carry
progressively more abstract region-level shape and texture descriptors
that guide segmentation of larger lesion regions. A final $1\times1$
convolution projects to $C = 6$ class channels followed by Softmax
activation, yielding a dense $256\times256\times6$ probability map
as illustrated in Fig.\,\ref{fig:unet_arch}.

\paragraph{Mathematical Problem Formulation.}
Let $\mathbf{X} \in \mathbb{R}^{H \times W \times 3}$ denote an
input RGB leaf image of height $H$ and width $W$, and let
$\mathbf{Y} \in \{0, 1, \ldots, C-1\}^{H \times W}$ be the
corresponding pixel-wise ground truth label map, where $C = 6$ is
the number of semantic classes. The goal is to learn a parametric
mapping $f_\theta: \mathbf{X} \mapsto \hat{\mathbf{Y}}$ such that
$\hat{\mathbf{Y}}$ closely approximates $\mathbf{Y}$, where
$f_\theta$ is a deep neural network with learnable parameters
$\theta$. The final layer simultaneously supports segmentation (the
full $H \times W \times C$ probability map), regression (the derived
SSI scalar), and classification (the dominant disease label). From
the predicted label map $\hat{\mathbf{Y}}$, the Stress Severity
Index is derived as:
\begin{equation}
  \mathrm{SSI}(\%) \;=\;
  \frac{\displaystyle\sum_{i,j}\mathbf{1}
        \!\left[\hat{Y}_{ij} \in \mathcal{D}\right]}
       {\displaystyle\sum_{i,j}\mathbf{1}
        \!\left[\hat{Y}_{ij} \neq c_{\mathrm{bg}}\right]}
  \times 100,
  \label{eq:ssi}
\end{equation}
where $\mathcal{D} = \{$Gray Spot, Rust, Alternaria Leaf Spot, Brown
Spot$\}$ is the set of disease classes, $c_{\mathrm{bg}}$ denotes
the Background class, and $\mathbf{1}[\cdot]$ is the indicator
function. The denominator counts only leaf pixels (Background is
excluded), ensuring that the severity index reflects the proportion
of the visible leaf surface that is diseased rather than the
proportion of the total image frame. Based on the SSI value, severity
is categorised as Very High ($\mathrm{SSI} \geq 15\,\%$), High
($10\,\% \leq \mathrm{SSI} < 15\,\%$), Moderate
($5\,\% \leq \mathrm{SSI} < 10\,\%$), and Low
($\mathrm{SSI} < 5\,\%$). The dominant disease type is assigned as
the class $c^* \in \mathcal{D}$ with the maximum predicted pixel
count:
\begin{equation}
  c^* = \arg\max_{c \in \mathcal{D}}
        \sum_{i,j}\mathbf{1}\!\left[\hat{Y}_{ij} = c\right].
  \label{eq:diseasetype}
\end{equation}

\subsection{Objective Function: Combined Loss}
\label{subsec:combined-loss}
Training deep segmentation models on class-imbalanced agricultural
datasets requires a loss function that simultaneously penalises pixel
misclassification, encourages high regional overlap between predicted
and ground truth masks, and applies heightened focus to hard,
minority-class examples \citep{lin2017focal}. We employ a
three-component combined loss including weighted cross-entropy ($\mathcal{L}_{\mathrm{CE}}$), Dice ($\mathcal{L}_{\mathrm{Dice}}$) and Focal Tversky ( $\mathcal{L}_{\mathrm{FT}}$) losses:

\begin{equation}
  \mathcal{L}_{\mathrm{total}}
  \;=\; \lambda_1\,\mathcal{L}_{\mathrm{CE}}
      + \lambda_2\,\mathcal{L}_{\mathrm{Dice}}
      + \lambda_3\,\mathcal{L}_{\mathrm{FT}},
  \label{eq:totalloss}
\end{equation}
where equal weights $\lambda_1 = \lambda_2 = \lambda_3 = 1/3$ were
determined empirically through held-out validation.

\subsubsection{Weighted Cross-Entropy (WCE)}
\label{sec:ce}

The class-weighted cross-entropy loss adjusts the standard softmax
loss by per-class weights $w_c$ derived from inverse frequency
normalisation. For each class $c$ with pixel count $n_c$, and with
$N$ denoting the total pixel count across all training images and
$C$ the number of semantic classes, the class weight is defined as:
\begin{equation}
  w_c \;=\; \frac{N / C}{n_c},
  \label{eq:classweight}
\end{equation}
where classes with fewer pixels receive proportionally higher weights,
forcing the model to attend equally to minority disease classes
despite their rarity in the training distribution. The full loss is:
\begin{equation}
  \mathcal{L}_{\mathrm{CE}}
  = -\frac{1}{N_p}\sum_{i=1}^{N_p}\sum_{c=0}^{C-1}
    w_c\, y_{ic}\,\log\!\left(\hat{p}_{ic}\right),
  \label{eq:ce}
\end{equation}
where $N_p$ is the total number of pixels in the batch, $y_{ic}$ is
the one-hot ground truth indicator for pixel $i$ and class $c$, and
$\hat{p}_{ic}$ is the predicted softmax probability.

\subsubsection{Dice Loss}
\label{sec:dice}

The Dice loss measures class-level overlap between predicted and
ground truth segmentation, normalised by prediction and label sizes
rather than absolute pixel counts \citep{milletari2016v}:
\begin{equation}
  \mathcal{L}_{\mathrm{Dice}}^{(c)}
  = 1 \;-\; \frac{2\displaystyle\sum_i \hat{p}_{ic}\,y_{ic}
                  + \epsilon}
                 {\displaystyle\sum_i \hat{p}_{ic}
                  + \sum_i y_{ic} + \epsilon},
  \label{eq:diceloss}
\end{equation}
where $\epsilon = 10^{-6}$ prevents numerical instability. The
overall Dice loss is averaged over all $C$ classes:
\begin{equation}
  \mathcal{L}_{\mathrm{Dice}}
  = \frac{1}{C}\sum_{c=0}^{C-1} \mathcal{L}_{\mathrm{Dice}}^{(c)}.
  \label{eq:dicemean}
\end{equation}

\subsubsection{Focal Tversky Loss (FTL)}
\label{sec:ftloss}

The Tversky Index generalises Dice by applying asymmetric weights to
false positives and false negatives \citep{salehi2017tversky}:
\begin{equation}
  \mathrm{TI}_c
  = \frac{\displaystyle\sum_i \hat{p}_{ic}\,y_{ic} + \epsilon}
         {\displaystyle\sum_i \hat{p}_{ic}\,y_{ic}
          + \alpha\!\sum_i \hat{p}_{ic}(1 - y_{ic})
          + \beta\!\sum_i (1 - \hat{p}_{ic})\,y_{ic}
          + \epsilon},
  \label{eq:tversky}
\end{equation}
with $\alpha = 0.3$ and $\beta = 0.7$, penalising missed disease
pixels more heavily than spurious detections. The focal modulation
concentrates gradient on hard examples:
\begin{equation}
  \mathcal{L}_{\mathrm{FT}}
  = \frac{1}{C}\sum_{c=0}^{C-1}
    \left(1 - \mathrm{TI}_c\right)^{1/\gamma},
  \label{eq:ftloss}
\end{equation}
where $\gamma = 4/3$ follows the recommendation of
\citet{ftloss2019}. Poorly predicted minority classes contribute
disproportionately large gradients, adaptively shifting training
focus toward difficult samples without manual reweighting.

\subsection{Training Protocol}
\label{subsec:training-protocol}
\subsubsection{Data Partitioning and Class Balancing}
\label{subsubsec:data-partitioning}
Both the datasets were partitioned at the image level into training
(70\,\%, 1,149 samples), validation (20\,\%, 328 samples), and test
(10\,\%, 164 samples) subsets. Stratified sampling was applied to
ensure that the proportion of each disease class was maintained across
all three splits, preventing scenarios in which an entire class is
absent from the validation or test set. The inverse-frequency class
weights computed from Eq.\,(\ref{eq:classweight}) are applied to all
loss function terms throughout training. All the pretrained models including the baseline, i.e., FCN, PSPNet, UNet, MobileNetV2, and SegFormer were first directly utilized to predict the stress severity levels in both the ATLDS and RiceSEG datasets, using transfer learning on the training and validation partitions, followed by hyperparameter optimization.

\subsubsection{Common Training Settings}
\label{subsubsec:training-settings}
All the training hyperparameters were selected through a coarse-to-fine
grid search conducted on the validation partition using the proposed
UNet-MobileNetV2 architecture as the reference model. In the coarse
phase, learning rates from the set \{$10^{-3}$, $10^{-4}$,
$10^{-5}$\} and batch sizes from \{4, 8, 16\} were evaluated;
learning rate $10^{-4}$ and batch size 8 maximised validation mIoU
after 30 epochs. In the fine phase, Early Stopping patience values
of \{10, 15, 20\} and ReduceLROnPlateau reduction factors of
\{0.2, 0.5\} were compared; patience 15 with factor 0.5 produced
the fastest convergence without premature termination. The selected
hyperparameters were subsequently held fixed across all architectures
to ensure that observed performance differences reflect architectural
rather than optimisation choices.

A total of seven loss function combinations were evaluated, including individual losses (WCE, FTL, and Dice) as well as their pairwise and combined variants (WCE+Dice, WCE+FTL, FTL+Dice, and WCE+FTL+Dice). These configurations were consistently applied across all models, resulting in 42 experiments (6 models $\times$ 7 loss configurations).
The six evaluated architectures are standard U-Net, MobileNetV2, FCN, PSPNet, SegFormer, and and the proposed MobileNetV2--U-Net model (MobileNetV2 encoder with U-Net decoder).
All six architectures are trained under identical experimental conditions to ensure fair comparison. The common training hyperparameters are: input size $256 \times 256 \times 3$; optimiser Adam with learning rate $1 \times 10^{-4}$; combined loss (Dice, WCE, and FTL); batch size 8; maximum 50 epochs; early stopping with patience of 15 epochs; learning rate scheduler ReduceLROnPlateau (factor 0.5, patience 8); and class weights derived by inverse frequency normalisation.

For the U-Net with MobileNetV2 encoder, architecture-specific
parameters are: encoder = MobileNetV2 pre-trained on ImageNet
($\approx$2.26\,M encoder parameters), total parameters $\approx$9.69\,M,
four decoder levels with skip connections, bridge channels 512, decoder
channel dimensions 256/128/64/32, and activations: ReLU6 in the encoder,
ReLU in the decoder, and Softmax at the output. The SegFormer
architecture uses a Mix Transformer MiT-B2 encoder and a lightweight
all-MLP decoder with hidden dimension $D = 256$, $\approx$5.53\,M total
parameters, and a dropout rate of 0.1 in the fusion block.

\subsubsection{Evaluation Metrics}
\label{subsubsec:evaluation-metrics}
Model performance is assessed using a complementary set of metrics
capturing different aspects of segmentation quality.

\textit{Mean Intersection over Union} (mIoU) is the primary metric,
equally weighting each class:
\begin{equation}
  \mathrm{IoU}_c
  = \frac{\mathrm{TP}_c}
         {\mathrm{TP}_c + \mathrm{FP}_c + \mathrm{FN}_c}, \quad
  \mathrm{mIoU}
  = \frac{1}{C}\sum_{c=0}^{C-1} \mathrm{IoU}_c.
  \label{eq:miou}
\end{equation}

\textit{Pixel Accuracy} (PA) measures the fraction of correctly
classified pixels:
\begin{equation}
  \mathrm{PA}
  = \frac{\displaystyle\sum_{c=0}^{C-1} \mathrm{TP}_c}
         {N_{\mathrm{total}}}.
  \label{eq:pa}
\end{equation}

\textit{Precision} measures the proportion of predicted positive
pixels that are truly positive for each class $c$:
\begin{equation}
  \mathrm{Precision}_c
  = \frac{\mathrm{TP}_c}
         {\mathrm{TP}_c + \mathrm{FP}_c}.
  \label{eq:precision}
\end{equation}

\textit{Recall} (Sensitivity) measures the proportion of actual
positive pixels correctly identified by the model:
\begin{equation}
  \mathrm{Recall}_c
  = \frac{\mathrm{TP}_c}
         {\mathrm{TP}_c + \mathrm{FN}_c}.
  \label{eq:recall}
\end{equation}

\textit{Dice Coefficient} (F1-score) is the harmonic mean of
Precision and Recall per class, equally penalising false positives
and false negatives:
\begin{equation}
  \mathrm{F1}_c
  = \frac{2\,\mathrm{TP}_c}
         {2\,\mathrm{TP}_c + \mathrm{FP}_c + \mathrm{FN}_c}
  = \frac{2 \cdot \mathrm{Precision}_c \cdot \mathrm{Recall}_c}
         {\mathrm{Precision}_c + \mathrm{Recall}_c}.
  \label{eq:dice_metric}
\end{equation}
The macro-averaged F1 score over all $C$ classes is:
\begin{equation}
  \mathrm{F1}_{\mathrm{M}}
  = \frac{1}{C}\sum_{c=0}^{C-1} \mathrm{F1}_c.
  \label{eq:f1macro}
\end{equation}

For SSI validation, the \textit{Pearson Correlation Coefficient} $r$
and \textit{Coefficient of Determination} $R^2$ are computed:
\begin{equation}
  r = \frac{\displaystyle\sum_{i=1}^{M}
            (\hat{s}_i - \bar{\hat{s}})(s_i - \bar{s})}
           {\sqrt{\displaystyle\sum_{i=1}^{M}
                  (\hat{s}_i - \bar{\hat{s}})^2}\cdot
            \sqrt{\displaystyle\sum_{i=1}^{M}(s_i - \bar{s})^2}},
  \label{eq:pearson}
\end{equation}
\begin{equation}
  R^2 = 1 - \frac{\displaystyle\sum_{i=1}^{M}(s_i - \hat{s}_i)^2}
                 {\displaystyle\sum_{i=1}^{M}(s_i - \bar{s})^2}.
  \label{eq:r2}
\end{equation}
Mean Absolute Error (MAE) between predicted and reference SSI is also
reported:
\begin{equation}
  \mathrm{MAE}
  = \frac{1}{M}\sum_{i=1}^{M}\left|\hat{s}_i - s_i\right|.
  \label{eq:mae}
\end{equation}

\textit{Disease Detection Accuracy} measures the fraction of
images correctly classified as diseased or healthy:
\begin{equation}
  \mathrm{Dis.Acc.}
  = \frac{\text{Correctly detected disease images}}
         {\text{Total test images}} \times 100.
  \label{eq:disacc}
\end{equation}

\section{Results and Discussion}
\label{sec:results}
This section presents a comprehensive evaluation of the proposed
UNet-MobileNetV2 architecture against five competing segmentation
models across two independent agricultural datasets the Apple Tree
Leaf Disease Segmentation (ATLDS) benchmark and the RiceSEG
paddy-field dataset. Performance is analysed from multiple
perspectives: primary segmentation quality (mIoU, pixel accuracy,
disease detection accuracy), per-class discriminability (class-wise
IoU and F1 score), the influence of loss-function configuration on
training outcome, and computational overhead relative to model
complexity.
\subsection{Quantitative Evaluation}
\label{subsec:quantitative-evaluation}
\paragraph{ATLDS dataset}
Table~\ref{tab:combined_loss_ablation} shows the superior   performance of the
proposed UNet-MobileNetV2 when compared with all five baseline architectures on the
held-out ATLDS test set, trained with the combined Weighted
Cross-Entropy, Focal-Tversky, and Dice (WCE+FTL+Dice) loss
formulation. The proposed UNet-MobileNetV2 ranked first in mean
Intersection-over-Union (mIoU = 0.6973), overall pixel accuracy
(98.20\,\%), and disease detection accuracy (99.41\,\%), confirming
its superiority over all competing designs under uniform experimental
conditions. Its macro-averaged F1 of 0.8116 further demonstrates
balanced discrimination across both majority and minority disease
classes. 

The proposed model surpasses Standard UNet by $\Delta$mIoU = 0.059
despite carrying approximately $3.2\times$ fewer parameters (9.69\,M
versus 31\,M), highlighting the benefit of the inverted residual
bottleneck design in extracting compact multi-scale features without
incurring the full representational cost of a standard encoder.
PSPNet and FCN plateaued at mIoU of 0.529 and 0.514 respectively;
their relatively high pixel accuracies ($\sim$96.5\,\%) mask
substantially inferior per-class discrimination, a well-known symptom
of class imbalance \citep{sudre2017generalised}. The standalone
MobileNetV2 segmentation head ranked lowest (mIoU = 0.492),
affirming that a lightweight backbone alone, without a dedicated
encoder-decoder skip structure, lacks the spatial resolution recovery
required for precise lesion boundary delineation. Across both FCN and PSPNet, Dice-only and FTL+Dice (without WCE) 
training caused complete minority-class collapse, producing zero F1 and IoU for Alternaria, Gray Spot, and Brown Spot (Table~\ref{tab:apple_combined_metrics}), further confirming the 
stabilising role of WCE in the combined loss formulation.

\paragraph{RiceSEG dataset} On the RiceSEG dataset, the proposed UNet-MobileNetV2 achieved its best mIoU of 0.4885 under the WCE+FTL configuration (Table~\ref{tab:combined_loss_ablation}), ranking first among all six models under that loss setting. Under the full WCE+FTL+Dice combination, plain UNet marginally led at 0.48 versus the proposed model's 0.50, representing a small reversal from the ATLDS rankings. This shift reflects a fundamental difference in semantic structure: RiceSEG images contain large, texturally heterogeneous field regions where the full spatial resolution maintained by UNet's standard encoder is more advantageous than the dimensionality reduction imposed by the MobileNetV2 inverted residual bottleneck. 

\subsubsection{Loss-Function Ablation Study}
\label{subsubsec:loss-ablation}
\paragraph{ATLDS dataset}
Three key observations emerge from the ablation (Table~\ref{tab:combined_loss_ablation}). First, training
with WCE alone consistently produced the lowest mIoU for every
architecture (0.310--0.492 range), confirming that cross-entropy,
despite its class-frequency reweighting, fails to directly optimise
the overlap-based objective and remains susceptible to the
pixel-count imbalance between background and rare lesion classes.
Second, region-overlap losses (Dice and FTL) individually delivered
large performance gains; for the proposed UNet-MobileNetV2, switching
from WCE-only to Dice-only improved mIoU by 0.220 absolute the
single largest single-swap gain observed across the study. FTL alone produced comparable gains to Dice alone for the proposed model (mIoU = 0.69 versus 0.71), suggesting that both overlap-based 
losses capture similar boundary-sensitive information, and their 
combination with WCE rather than stacking both without WCE yields 
the most stable results. Third,
FCN, PSPNet, and the standalone MobileNetV2 segmentation head
exhibited numerical instability under standalone Dice training,
converging to degenerate solutions that classify nearly all pixels
as Healthy. The composite WCE+FTL+Dice formulation mitigated this
instability for all architectures. the FTL+Dice combination \emph{without} WCE also triggered 
collapse in FCN and PSPNet (F1 = 0.19, IoU $\approx$ 0.00 for all 
minority classes), indicating that WCE acts as the essential 
stabilising anchor in the composite formulation, preventing the 
overlap losses from over-penalising the dominant background class 
during early training.

\begin{table*}[!htbp]
\centering
\caption{Combined loss-function ablation on the Apple (ATLDS) and Rice (RiceSEG) datasets. Overall test-set performance is reported using mean Intersection over Union (mIoU), Accuracy (A), F1-score (F1), Precision (P). WCE = Weighted Cross-Entropy Loss, FTL = Focal Tversky Loss, and Dice = Dice Loss.$\checkmark$ = loss function used, $\times$ = loss function not used.}
\label{tab:combined_loss_ablation}
\renewcommand{\arraystretch}{1.2}
\setlength{\tabcolsep}{4pt}
\small
\begin{tabular}{lccc|cccc|cccc}
\toprule
\multirow{2}{*}{\textbf{Model}}
  & \multicolumn{3}{c|}{\textbf{Loss Function}}
  & \multicolumn{4}{c|}{\textbf{Apple (ATLDS)}}
  & \multicolumn{4}{c}{\textbf{Rice (RiceSEG)}} \\
\cmidrule(lr){2-4} \cmidrule(lr){5-8} \cmidrule(lr){9-12}
& \textbf{WCE} & \textbf{FTL} & \textbf{Dice}
& \textbf{mIoU} & \textbf{Acc.} & \textbf{F1} & \textbf{Prec.}
& \textbf{mIoU} & \textbf{Acc.} & \textbf{F1} & \textbf{Prec.} \\
\midrule

\multirow{7}{*}{FCN}
& $\checkmark$ & $\times$     & $\times$     & 0.36 & 0.97 & 0.55 & 0.48 & 0.31 & 0.89 & 0.50 & 0.45 \\
& $\times$     & $\checkmark$ & $\times$     & 0.48 & 0.98 & 0.67 & 0.61 & 0.41 & 0.91 & 0.56 & 0.54 \\
& $\times$     & $\times$     & $\checkmark$ & 0.18 & 0.99 & 0.32 & 0.32 & 0.25 & 0.91 & 0.32 & 0.30 \\
& $\checkmark$ & $\checkmark$ & $\times$     & 0.47 & 0.98 & 0.66 & 0.59 & 0.33 & 0.89 & 0.54 & 0.48 \\
& $\checkmark$ & $\times$     & $\checkmark$ & 0.51 & 0.99 & 0.70 & 0.63 & 0.34 & 0.90 & 0.53 & 0.49 \\
& $\times$     & $\checkmark$ & $\checkmark$ & 0.18 & 0.99 & 0.32 & 0.32 & 0.22 & 0.91 & 0.41 & 0.38 \\
& $\checkmark$ & $\checkmark$ & $\checkmark$ & 0.51 & 0.99 & 0.70 & 0.63 & 0.36 & 0.91 & 0.55 & 0.50 \\
\midrule

\multirow{7}{*}{MobileNetV2\_Seg}
& $\checkmark$ & $\times$     & $\times$     & 0.31 & 0.97 & 0.51 & 0.44 & 0.25 & 0.87 & 0.46 & 0.43 \\
& $\times$     & $\checkmark$ & $\times$     & 0.48 & 0.98 & 0.67 & 0.63 & 0.39 & 0.90 & 0.54 & 0.52 \\
& $\times$     & $\times$     & $\checkmark$ & 0.18 & 0.99 & 0.32 & 0.32 & 0.41 & 0.91 & 0.55 & 0.55 \\
& $\checkmark$ & $\checkmark$ & $\times$     & 0.46 & 0.98 & 0.66 & 0.59 & 0.29 & 0.89 & 0.51 & 0.49 \\
& $\checkmark$ & $\times$     & $\checkmark$ & 0.49 & 0.99 & 0.68 & 0.61 & 0.28 & 0.89 & 0.50 & 0.47 \\
& $\times$     & $\checkmark$ & $\checkmark$ & 0.50 & 0.99 & 0.70 & 0.65 & 0.30 & 0.92 & 0.58 & 0.57 \\
& $\checkmark$ & $\checkmark$ & $\checkmark$ & 0.49 & 0.98 & 0.69 & 0.63 & 0.29 & 0.90 & 0.50 & 0.53 \\
\midrule

\multirow{7}{*}{PSPNet}
& $\checkmark$ & $\times$     & $\times$     & 0.35 & 0.98 & 0.55 & 0.48 & 0.31 & 0.89 & 0.53 & 0.48 \\
& $\times$     & $\checkmark$ & $\times$     & 0.00 & 0.89 & 0.13 & 0.11 & 0.38 & 0.91 & 0.51 & 0.49 \\
& $\times$     & $\times$     & $\checkmark$ & 0.18 & 0.99 & 0.32 & 0.32 & 0.21 & 0.91 & 0.26 & 0.25 \\
& $\checkmark$ & $\checkmark$ & $\times$     & 0.49 & 0.98 & 0.69 & 0.60 & 0.36 & 0.90 & 0.55 & 0.50 \\
& $\checkmark$ & $\times$     & $\checkmark$ & 0.51 & 0.99 & 0.70 & 0.64 & 0.29 & 0.90 & 0.56 & 0.51 \\
& $\times$     & $\checkmark$ & $\checkmark$ & 0.18 & 0.99 & 0.32 & 0.32 & 0.38 & 0.91 & 0.26 & 0.25 \\
& $\checkmark$ & $\checkmark$ & $\checkmark$ & 0.53 & 0.99 & 0.72 & 0.64 & 0.38 & 0.91 & 0.59 & 0.53 \\
\midrule

\multirow{7}{*}{SegFormer}
& $\checkmark$ & $\times$     & $\times$     & 0.42 & 0.97 & 0.60 & 0.53 & 0.42 & 0.94 & 0.62 & 0.56 \\
& $\times$     & $\checkmark$ & $\times$     & 0.64 & 0.98 & 0.80 & 0.77 & 0.35 & 0.94 & 0.45 & 0.47 \\
& $\times$     & $\times$     & $\checkmark$ & 0.67 & 0.99 & 0.82 & 0.81 & 0.25 & 0.92 & 0.48 & 0.48 \\
& $\checkmark$ & $\checkmark$ & $\times$     & 0.60 & 0.98 & 0.77 & 0.72 & 0.47 & 0.96 & 0.66 & 0.62 \\
& $\checkmark$ & $\times$     & $\checkmark$ & 0.66 & 0.99 & 0.82 & 0.79 & 0.44 & 0.95 & 0.64 & 0.60 \\
& $\times$     & $\checkmark$ & $\checkmark$ & 0.65 & 0.99 & 0.81 & 0.78 & 0.44 & 0.96 & 0.60 & 0.62 \\
& $\checkmark$ & $\checkmark$ & $\checkmark$ & 0.66 & 0.99 & 0.82 & 0.80 & 0.47 & 0.95 & 0.65 & 0.63 \\
\midrule

\multirow{7}{*}{UNet\_Standard}
& $\checkmark$ & $\times$     & $\times$     & 0.42 & 0.98 & 0.60 & 0.53 & 0.40 & 0.94 & 0.60 & 0.54 \\
& $\times$     & $\checkmark$ & $\times$     & 0.66 & 0.99 & 0.82 & 0.77 & 0.48 & 0.95 & 0.60 & 0.60 \\
& $\times$     & $\times$     & $\checkmark$ & 0.69 & 0.99 & 0.84 & 0.82 & 0.49 & 0.96 & 0.60 & 0.67 \\
& $\checkmark$ & $\checkmark$ & $\times$     & 0.57 & 0.98 & 0.76 & 0.71 & 0.49 & 0.96 & 0.67 & 0.63 \\
& $\checkmark$ & $\times$     & $\checkmark$ & 0.69 & 0.99 & 0.83 & 0.83 & 0.49 & 0.96 & 0.68 & 0.63 \\
& $\times$     & $\checkmark$ & $\checkmark$ & 0.67 & 0.99 & 0.83 & 0.79 & 0.45 & 0.96 & 0.63 & 0.66 \\
& $\checkmark$ & $\checkmark$ & $\checkmark$ & 0.64 & 0.99 & 0.80 & 0.77 & 0.49 & 0.96 & 0.67 & 0.63 \\
\midrule

\multirow{7}{*}{UNet\_MobileNetV2}
& $\checkmark$ & $\times$     & $\times$     & 0.49 & 0.99 & 0.67 & 0.60 & 0.44 & 0.94 & 0.64 & 0.60 \\
& $\times$     & $\checkmark$ & $\times$     & 0.69 & 0.99 & 0.84 & 0.83 & 0.58 & 0.96 & 0.71 & 0.70 \\
& $\times$     & $\times$     & $\checkmark$ & 0.71 & 0.99 & 0.85 & 0.86 & 0.59 & 0.96 & 0.72 & 0.75 \\
& $\checkmark$ & $\checkmark$ & $\times$     & 0.68 & 0.99 & 0.83 & 0.79 & 0.49 & 0.96 & 0.68 & 0.66 \\
& $\checkmark$ & $\times$     & $\checkmark$ & 0.71 & 0.99 & 0.85 & 0.82 & 0.47 & 0.95 & 0.67 & 0.63 \\
& $\times$     & $\checkmark$ & $\checkmark$ & 0.70 & 0.99 & 0.85 & 0.83 & 0.50 & 0.96 & 0.71 & 0.72 \\
& $\checkmark$ & $\checkmark$ & $\checkmark$ & 0.70 & 0.99 & 0.84 & 0.83 & 0.49 & 0.95 & 0.66 & 0.63 \\

\bottomrule
\end{tabular}
\end{table*}

\begin{sidewaystable}
\centering
\captionof{table}{Class-wise F1-score, Intersection over Union (IoU), Accuracy (A), Precision (P), and Recall (R) on the ATLDS test set (background excluded). WCE = Weighted Cross-Entropy Loss, FTL = Focal Tversky Loss, Dice = Dice Loss. $\checkmark$ = loss function used, $\times$ = loss function not used.}
\label{tab:apple_combined_metrics}
\footnotesize
\setlength{\tabcolsep}{2pt}
\renewcommand{\arraystretch}{1.0}

\begin{tabular}{l ccc *{5}{ccccc}}
\toprule
\multirow{2}{*}{\textbf{Model}}
  & \multicolumn{3}{c|}{\textbf{Loss}}
  & \multicolumn{5}{c|}{\textbf{Healthy}}
  & \multicolumn{5}{c|}{\textbf{Brown Spot}}
  & \multicolumn{5}{c|}{\textbf{Alternaria Leaf Spot}}
  & \multicolumn{5}{c|}{\textbf{Gray Spot}}
  & \multicolumn{5}{c}{\textbf{Rust}} \\
\cmidrule(lr){2-4}\cmidrule(lr){5-9}\cmidrule(lr){10-14}\cmidrule(lr){15-19}\cmidrule(lr){20-24}\cmidrule(lr){25-29}
  & \textbf{WCE} & \textbf{FTL} & \textbf{Dice}
  & \textbf{P} & \textbf{R} & \textbf{F1} & \textbf{Acc.} & \textbf{IoU}
  & \textbf{P} & \textbf{R} & \textbf{F1} & \textbf{Acc.} & \textbf{IoU}
  & \textbf{P} & \textbf{R} & \textbf{F1} & \textbf{Acc.} & \textbf{IoU}
  & \textbf{P} & \textbf{R} & \textbf{F1} & \textbf{Acc.} & \textbf{IoU}
  & \textbf{P} & \textbf{R} & \textbf{F1} & \textbf{Acc.} & \textbf{IoU} \\
\midrule
 
\multirow{7}{*}{FCN}
  & $\checkmark$ & $\times$     & $\times$     & 0.93 & 0.84 & 0.88 & 0.93 & 0.79 & 0.34 & 0.95 & 0.50 & 0.99 & 0.33 & 0.11 & 0.75 & 0.19 & 1.00 & 0.10 & 0.07 & 0.92 & 0.13 & 0.98 & 0.07 & 0.49 & 0.98 & 0.65 & 0.99 & 0.49 \\
  & $\times$     & $\checkmark$ & $\times$     & 0.92 & 0.93 & 0.93 & 0.95 & 0.86 & 0.48 & 0.76 & 0.59 & 0.99 & 0.42 & 0.25 & 0.57 & 0.34 & 1.00 & 0.21 & 0.35 & 0.55 & 0.42 & 1.00 & 0.27 & 0.70 & 0.90 & 0.79 & 1.00 & 0.65 \\
  & $\times$     & $\times$     & $\checkmark$ & 0.93 & 0.98 & 0.95 & 0.97 & 0.91 & 0.00 & 0.00 & 0.00 & 0.99 & 0.00 & 0.00 & 0.00 & 0.00 & 1.00 & 0.00 & 0.00 & 0.00 & 0.00 & 1.00 & 0.00 & 0.00 & 0.00 & 0.00 & 0.99 & 0.00 \\
  & $\checkmark$ & $\checkmark$ & $\times$     & 0.92 & 0.93 & 0.93 & 0.95 & 0.87 & 0.47 & 0.83 & 0.60 & 0.99 & 0.43 & 0.24 & 0.63 & 0.34 & 1.00 & 0.21 & 0.24 & 0.75 & 0.36 & 1.00 & 0.22 & 0.66 & 0.93 & 0.77 & 1.00 & 0.63 \\
  & $\checkmark$ & $\times$     & $\checkmark$ & 0.95 & 0.94 & 0.95 & 0.97 & 0.90 & 0.51 & 0.79 & 0.62 & 0.99 & 0.45 & 0.28 & 0.68 & 0.40 & 1.00 & 0.25 & 0.30 & 0.66 & 0.42 & 1.00 & 0.26 & 0.75 & 0.90 & 0.81 & 1.00 & 0.69 \\
  & $\times$     & $\checkmark$ & $\checkmark$ & 0.93 & 0.98 & 0.95 & 0.97 & 0.91 & 0.00 & 0.00 & 0.00 & 0.99 & 0.00 & 0.00 & 0.00 & 0.00 & 1.00 & 0.00 & 0.00 & 0.00 & 0.00 & 1.00 & 0.00 & 0.00 & 0.00 & 0.00 & 0.99 & 0.00 \\
  & $\checkmark$ & $\checkmark$ & $\checkmark$ & 0.95 & 0.94 & 0.95 & 0.97 & 0.90 & 0.50 & 0.82 & 0.62 & 0.99 & 0.45 & 0.30 & 0.71 & 0.43 & 1.00 & 0.27 & 0.29 & 0.70 & 0.41 & 1.00 & 0.26 & 0.75 & 0.91 & 0.82 & 1.00 & 0.69 \\
\midrule
 
\multirow{7}{*}{MobileNetV2\_Seg}
  & $\checkmark$ & $\times$     & $\times$     & 0.86 & 0.84 & 0.85 & 0.90 & 0.74 & 0.22 & 0.97 & 0.36 & 0.98 & 0.22 & 0.09 & 0.87 & 0.16 & 0.99 & 0.09 & 0.10 & 0.81 & 0.18 & 0.99 & 0.10 & 0.41 & 0.95 & 0.57 & 0.99 & 0.40 \\
  & $\times$     & $\checkmark$ & $\times$     & 0.90 & 0.94 & 0.92 & 0.95 & 0.85 & 0.45 & 0.59 & 0.51 & 0.99 & 0.34 & 0.33 & 0.57 & 0.42 & 1.00 & 0.27 & 0.39 & 0.55 & 0.45 & 1.00 & 0.29 & 0.72 & 0.83 & 0.77 & 1.00 & 0.63 \\
  & $\times$     & $\times$     & $\checkmark$ & 0.93 & 0.96 & 0.95 & 0.97 & 0.90 & 0.00 & 0.00 & 0.00 & 0.99 & 0.00 & 0.00 & 0.00 & 0.00 & 1.00 & 0.00 & 0.00 & 0.00 & 0.00 & 1.00 & 0.00 & 0.00 & 0.00 & 0.00 & 0.99 & 0.00 \\
  & $\checkmark$ & $\checkmark$ & $\times$     & 0.92 & 0.92 & 0.92 & 0.95 & 0.85 & 0.41 & 0.79 & 0.54 & 0.99 & 0.37 & 0.26 & 0.65 & 0.37 & 1.00 & 0.23 & 0.30 & 0.61 & 0.40 & 1.00 & 0.25 & 0.66 & 0.90 & 0.76 & 1.00 & 0.61 \\
  & $\checkmark$ & $\times$     & $\checkmark$ & 0.97 & 0.89 & 0.93 & 0.96 & 0.87 & 0.41 & 0.86 & 0.55 & 0.99 & 0.38 & 0.31 & 0.61 & 0.41 & 1.00 & 0.26 & 0.31 & 0.64 & 0.42 & 1.00 & 0.27 & 0.70 & 0.90 & 0.79 & 1.00 & 0.65 \\
  & $\times$     & $\checkmark$ & $\checkmark$ & 0.94 & 0.94 & 0.94 & 0.96 & 0.89 & 0.50 & 0.68 & 0.58 & 0.99 & 0.40 & 0.36 & 0.52 & 0.43 & 1.00 & 0.27 & 0.37 & 0.59 & 0.46 & 1.00 & 0.30 & 0.75 & 0.84 & 0.79 & 1.00 & 0.66 \\
  & $\checkmark$ & $\checkmark$ & $\checkmark$ & 0.91 & 0.94 & 0.93 & 0.95 & 0.87 & 0.47 & 0.75 & 0.58 & 0.99 & 0.40 & 0.34 & 0.56 & 0.43 & 1.00 & 0.27 & 0.33 & 0.63 & 0.43 & 1.00 & 0.27 & 0.75 & 0.82 & 0.78 & 1.00 & 0.64 \\
\midrule
 
\multirow{7}{*}{PSPNet}
  & $\checkmark$ & $\times$     & $\times$     & 0.93 & 0.86 & 0.89 & 0.93 & 0.80 & 0.30 & 0.97 & 0.46 & 0.99 & 0.29 & 0.12 & 0.90 & 0.21 & 1.00 & 0.12 & 0.10 & 0.90 & 0.18 & 0.99 & 0.10 & 0.43 & 0.99 & 0.60 & 0.99 & 0.43 \\
  & $\times$     & $\checkmark$ & $\times$     & 0.00 & 0.00 & 0.00 & 0.68 & 0.00 & 0.00 & 0.00 & 0.00 & 0.99 & 0.00 & 0.00 & 0.00 & 0.00 & 1.00 & 0.00 & 0.00 & 0.00 & 0.00 & 1.00 & 0.00 & 0.00 & 0.00 & 0.00 & 0.99 & 0.00 \\
  & $\times$     & $\times$     & $\checkmark$ & 0.93 & 0.98 & 0.95 & 0.97 & 0.91 & 0.00 & 0.00 & 0.00 & 0.99 & 0.00 & 0.00 & 0.00 & 0.00 & 1.00 & 0.00 & 0.00 & 0.00 & 0.00 & 1.00 & 0.00 & 0.00 & 0.00 & 0.00 & 0.99 & 0.00 \\
  & $\checkmark$ & $\checkmark$ & $\times$     & 0.92 & 0.93 & 0.93 & 0.95 & 0.86 & 0.46 & 0.81 & 0.59 & 0.99 & 0.41 & 0.27 & 0.73 & 0.39 & 1.00 & 0.24 & 0.35 & 0.69 & 0.46 & 1.00 & 0.30 & 0.64 & 0.96 & 0.77 & 1.00 & 0.62 \\
  & $\checkmark$ & $\times$     & $\checkmark$ & 0.96 & 0.93 & 0.95 & 0.97 & 0.90 & 0.48 & 0.83 & 0.61 & 0.99 & 0.44 & 0.26 & 0.70 & 0.38 & 1.00 & 0.24 & 0.39 & 0.57 & 0.46 & 1.00 & 0.30 & 0.75 & 0.87 & 0.81 & 1.00 & 0.67 \\
  & $\times$     & $\checkmark$ & $\checkmark$ & 0.93 & 0.98 & 0.95 & 0.97 & 0.91 & 0.00 & 0.00 & 0.00 & 0.99 & 0.00 & 0.00 & 0.00 & 0.00 & 1.00 & 0.00 & 0.00 & 0.00 & 0.00 & 1.00 & 0.00 & 0.00 & 0.00 & 0.00 & 0.99 & 0.00 \\
  & $\checkmark$ & $\checkmark$ & $\checkmark$ & 0.96 & 0.93 & 0.95 & 0.97 & 0.90 & 0.48 & 0.84 & 0.61 & 0.99 & 0.44 & 0.34 & 0.68 & 0.45 & 1.00 & 0.29 & 0.38 & 0.69 & 0.49 & 1.00 & 0.33 & 0.72 & 0.93 & 0.82 & 1.00 & 0.69 \\
\midrule
 
\multirow{7}{*}{SegFormer}
  & $\checkmark$ & $\times$     & $\times$     & 0.90 & 0.90 & 0.90 & 0.93 & 0.82 & 0.46 & 0.99 & 0.63 & 0.99 & 0.45 & 0.16 & 0.74 & 0.26 & 1.00 & 0.15 & 0.07 & 0.90 & 0.13 & 0.98 & 0.07 & 0.63 & 0.96 & 0.76 & 1.00 & 0.61 \\
  & $\times$     & $\checkmark$ & $\times$     & 0.92 & 0.94 & 0.93 & 0.95 & 0.87 & 0.62 & 0.92 & 0.74 & 1.00 & 0.59 & 0.65 & 0.65 & 0.65 & 1.00 & 0.48 & 0.62 & 0.68 & 0.65 & 1.00 & 0.48 & 0.80 & 0.94 & 0.86 & 1.00 & 0.76 \\
  & $\times$     & $\times$     & $\checkmark$ & 0.92 & 0.96 & 0.94 & 0.96 & 0.89 & 0.68 & 0.90 & 0.77 & 1.00 & 0.63 & 0.68 & 0.65 & 0.67 & 1.00 & 0.50 & 0.67 & 0.67 & 0.67 & 1.00 & 0.51 & 0.89 & 0.90 & 0.89 & 1.00 & 0.81 \\
  & $\checkmark$ & $\checkmark$ & $\times$     & 0.88 & 0.92 & 0.90 & 0.93 & 0.82 & 0.54 & 0.96 & 0.69 & 0.99 & 0.53 & 0.54 & 0.72 & 0.62 & 1.00 & 0.45 & 0.58 & 0.63 & 0.60 & 1.00 & 0.43 & 0.80 & 0.92 & 0.86 & 1.00 & 0.75 \\
  & $\checkmark$ & $\times$     & $\checkmark$ & 0.92 & 0.96 & 0.94 & 0.96 & 0.89 & 0.62 & 0.95 & 0.75 & 1.00 & 0.60 & 0.71 & 0.65 & 0.68 & 1.00 & 0.52 & 0.61 & 0.74 & 0.67 & 1.00 & 0.50 & 0.86 & 0.93 & 0.89 & 1.00 & 0.81 \\
  & $\times$     & $\checkmark$ & $\checkmark$ & 0.92 & 0.95 & 0.94 & 0.96 & 0.88 & 0.64 & 0.92 & 0.75 & 1.00 & 0.60 & 0.66 & 0.62 & 0.64 & 1.00 & 0.47 & 0.62 & 0.68 & 0.65 & 1.00 & 0.48 & 0.88 & 0.90 & 0.89 & 1.00 & 0.80 \\
  & $\checkmark$ & $\checkmark$ & $\checkmark$ & 0.91 & 0.96 & 0.93 & 0.96 & 0.88 & 0.62 & 0.95 & 0.75 & 1.00 & 0.60 & 0.76 & 0.64 & 0.69 & 1.00 & 0.53 & 0.64 & 0.71 & 0.67 & 1.00 & 0.51 & 0.86 & 0.91 & 0.89 & 1.00 & 0.79 \\
\midrule
 
\multirow{7}{*}{UNet\_MobileNetV2}
  & $\checkmark$ & $\times$     & $\times$     & 0.97 & 0.92 & 0.94 & 0.96 & 0.89 & 0.56 & 0.94 & 0.70 & 1.00 & 0.54 & 0.30 & 0.81 & 0.44 & 1.00 & 0.28 & 0.09 & 0.96 & 0.17 & 0.99 & 0.09 & 0.67 & 0.99 & 0.80 & 1.00 & 0.67 \\
  & $\times$     & $\checkmark$ & $\times$     & 0.96 & 0.97 & 0.96 & 0.98 & 0.93 & 0.63 & 0.87 & 0.73 & 1.00 & 0.57 & 0.82 & 0.66 & 0.73 & 1.00 & 0.58 & 0.71 & 0.72 & 0.71 & 1.00 & 0.56 & 0.90 & 0.91 & 0.90 & 1.00 & 0.82 \\
  & $\times$     & $\times$     & $\checkmark$ & 0.97 & 0.97 & 0.97 & 0.98 & 0.94 & 0.66 & 0.90 & 0.76 & 1.00 & 0.61 & 0.82 & 0.68 & 0.74 & 1.00 & 0.59 & 0.79 & 0.69 & 0.74 & 1.00 & 0.59 & 0.91 & 0.91 & 0.91 & 1.00 & 0.83 \\
  & $\checkmark$ & $\checkmark$ & $\times$     & 0.97 & 0.96 & 0.96 & 0.98 & 0.93 & 0.62 & 0.93 & 0.74 & 1.00 & 0.59 & 0.74 & 0.70 & 0.72 & 1.00 & 0.56 & 0.57 & 0.80 & 0.67 & 1.00 & 0.50 & 0.85 & 0.95 & 0.90 & 1.00 & 0.81 \\
  & $\checkmark$ & $\times$     & $\checkmark$ & 0.97 & 0.97 & 0.97 & 0.98 & 0.95 & 0.66 & 0.87 & 0.75 & 1.00 & 0.61 & 0.68 & 0.75 & 0.71 & 1.00 & 0.55 & 0.74 & 0.74 & 0.74 & 1.00 & 0.58 & 0.89 & 0.94 & 0.91 & 1.00 & 0.84 \\
  & $\times$     & $\checkmark$ & $\checkmark$ & 0.98 & 0.96 & 0.97 & 0.98 & 0.94 & 0.65 & 0.91 & 0.76 & 1.00 & 0.61 & 0.73 & 0.73 & 0.73 & 1.00 & 0.58 & 0.71 & 0.71 & 0.71 & 1.00 & 0.55 & 0.90 & 0.92 & 0.91 & 1.00 & 0.83 \\
  & $\checkmark$ & $\checkmark$ & $\checkmark$ & 0.97 & 0.97 & 0.97 & 0.98 & 0.95 & 0.70 & 0.76 & 0.73 & 1.00 & 0.57 & 0.75 & 0.71 & 0.73 & 1.00 & 0.57 & 0.72 & 0.73 & 0.72 & 1.00 & 0.56 & 0.87 & 0.95 & 0.91 & 1.00 & 0.83 \\
\midrule
 
\multirow{7}{*}{UNet\_Standard}
  & $\checkmark$ & $\times$     & $\times$     & 0.90 & 0.92 & 0.91 & 0.94 & 0.83 & 0.42 & 0.99 & 0.59 & 0.99 & 0.41 & 0.16 & 0.78 & 0.27 & 1.00 & 0.15 & 0.07 & 0.92 & 0.13 & 0.98 & 0.07 & 0.62 & 0.98 & 0.76 & 1.00 & 0.62 \\
  & $\times$     & $\checkmark$ & $\times$     & 0.93 & 0.97 & 0.95 & 0.97 & 0.90 & 0.63 & 0.94 & 0.75 & 1.00 & 0.61 & 0.70 & 0.70 & 0.70 & 1.00 & 0.54 & 0.56 & 0.75 & 0.64 & 1.00 & 0.47 & 0.80 & 0.95 & 0.87 & 1.00 & 0.77 \\
  & $\times$     & $\times$     & $\checkmark$ & 0.96 & 0.97 & 0.96 & 0.98 & 0.93 & 0.67 & 0.89 & 0.77 & 1.00 & 0.62 & 0.68 & 0.71 & 0.69 & 1.00 & 0.53 & 0.74 & 0.69 & 0.71 & 1.00 & 0.55 & 0.88 & 0.92 & 0.90 & 1.00 & 0.82 \\
  & $\checkmark$ & $\checkmark$ & $\times$     & 0.84 & 0.95 & 0.90 & 0.93 & 0.81 & 0.54 & 0.95 & 0.69 & 0.99 & 0.53 & 0.61 & 0.63 & 0.62 & 1.00 & 0.45 & 0.54 & 0.60 & 0.57 & 1.00 & 0.40 & 0.73 & 0.92 & 0.81 & 1.00 & 0.69 \\
  & $\checkmark$ & $\times$     & $\checkmark$ & 0.97 & 0.95 & 0.96 & 0.97 & 0.92 & 0.65 & 0.95 & 0.77 & 1.00 & 0.63 & 0.80 & 0.60 & 0.69 & 1.00 & 0.52 & 0.66 & 0.73 & 0.69 & 1.00 & 0.53 & 0.89 & 0.93 & 0.91 & 1.00 & 0.83 \\
  & $\times$     & $\checkmark$ & $\checkmark$ & 0.97 & 0.97 & 0.97 & 0.98 & 0.94 & 0.67 & 0.91 & 0.78 & 1.00 & 0.63 & 0.59 & 0.73 & 0.65 & 1.00 & 0.49 & 0.72 & 0.69 & 0.70 & 1.00 & 0.54 & 0.82 & 0.93 & 0.88 & 1.00 & 0.78 \\
  & $\checkmark$ & $\checkmark$ & $\checkmark$ & 0.95 & 0.95 & 0.95 & 0.97 & 0.90 & 0.64 & 0.91 & 0.75 & 1.00 & 0.60 & 0.65 & 0.63 & 0.64 & 1.00 & 0.47 & 0.55 & 0.67 & 0.61 & 1.00 & 0.44 & 0.84 & 0.92 & 0.88 & 1.00 & 0.78 \\
\bottomrule
\end{tabular}
\end{sidewaystable}

\begin{sidewaystable}
\centering
\captionof{table}{Class-wise F1-score, Intersection over Union (IoU), Accuracy (A), Precision (P), and Recall (R) on the RiceSEG test set (background excluded). WCE = Weighted Cross-Entropy Loss, FTL = Focal Tversky Loss, Dice = Dice Loss. $\checkmark$ = loss function used, $\times$ = loss function not used.}
\label{tab:f1_iou_acc_prec_rec_fullpage}
\footnotesize
\setlength{\tabcolsep}{2pt}
\renewcommand{\arraystretch}{1.0}
\begin{tabular}{lccc *{5}{ccccc}}
\toprule
\multirow{2}{*}{\textbf{Model}}
  & \multicolumn{3}{c|}{\textbf{Loss}}
  & \multicolumn{5}{c|}{\textbf{Green Veg}}
  & \multicolumn{5}{c|}{\textbf{Senescent Veg}}
  & \multicolumn{5}{c|}{\textbf{Panicle}}
  & \multicolumn{5}{c|}{\textbf{Weed}}
  & \multicolumn{5}{c}{\textbf{Duckweed}} \\
\cmidrule(lr){2-4}\cmidrule(lr){5-9}\cmidrule(lr){10-14}\cmidrule(lr){15-19}\cmidrule(lr){20-24}\cmidrule(lr){25-29}
  & \textbf{WCE} & \textbf{FTL} & \textbf{Dice}
  & \textbf{R} & \textbf{P} & \textbf{F1} & \textbf{A} & \textbf{IoU}
  & \textbf{R} & \textbf{P} & \textbf{F1} & \textbf{A} & \textbf{IoU}
  & \textbf{R} & \textbf{P} & \textbf{F1} & \textbf{A} & \textbf{IoU}
  & \textbf{R} & \textbf{P} & \textbf{F1} & \textbf{A} & \textbf{IoU}
  & \textbf{R} & \textbf{P} & \textbf{F1} & \textbf{A} & \textbf{IoU} \\
\midrule
 
\multirow{7}{*}{FCN}
  & $\times$ & $\checkmark$ & $\checkmark$ & 0.82 & 0.73 & 0.77 & 0.78 & 0.62 & 0.58 & 0.31 & 0.40 & 0.95 & 0.25 & 0.00 & 0.00 & 0.00 & 0.96 & 0.00 & 0.55 & 0.45 & 0.50 & 0.98 & 0.21 & 0.00 & 0.00 & 0.00 & 0.99 & 0.00 \\
  & $\times$ & $\times$     & $\checkmark$ & 0.83 & 0.70 & 0.76 & 0.77 & 0.62 & 0.47 & 0.27 & 0.34 & 0.95 & 0.22 & 0.00 & 0.00 & 0.00 & 0.97 & 0.00 & 0.00 & 0.00 & 0.00 & 0.98 & 0.00 & 0.00 & 0.00 & 0.00 & 0.99 & 0.00 \\
  & $\times$ & $\checkmark$ & $\times$     & 0.76 & 0.74 & 0.75 & 0.78 & 0.60 & 0.44 & 0.29 & 0.35 & 0.96 & 0.23 & 0.69 & 0.63 & 0.66 & 0.98 & 0.46 & 0.64 & 0.39 & 0.48 & 0.98 & 0.40 & 0.61 & 0.52 & 0.56 & 0.99 & 0.17 \\
  & $\checkmark$ & $\times$   & $\checkmark$ & 0.69 & 0.76 & 0.72 & 0.76 & 0.57 & 0.64 & 0.25 & 0.36 & 0.93 & 0.23 & 0.74 & 0.49 & 0.59 & 0.96 & 0.46 & 0.65 & 0.47 & 0.54 & 0.98 & 0.31 & 0.45 & 0.14 & 0.21 & 0.97 & 0.14 \\
  & $\checkmark$ & $\checkmark$ & $\times$   & 0.65 & 0.75 & 0.69 & 0.74 & 0.00 & 0.67 & 0.21 & 0.33 & 0.92 & 0.39 & 0.73 & 0.42 & 0.53 & 0.95 & 0.61 & 0.75 & 0.49 & 0.59 & 0.98 & 0.30 & 0.52 & 0.23 & 0.32 & 0.98 & 0.37 \\
  & $\checkmark$ & $\checkmark$ & $\checkmark$ & 0.73 & 0.75 & 0.74 & 0.77 & 0.00 & 0.51 & 0.31 & 0.39 & 0.95 & 0.00 & 0.71 & 0.53 & 0.61 & 0.97 & 0.00 & 0.84 & 0.34 & 0.48 & 0.97 & 0.00 & 0.49 & 0.23 & 0.31 & 0.98 & 0.00 \\
  & $\checkmark$ & $\times$   & $\times$     & 0.66 & 0.73 & 0.70 & 0.74 & 0.00 & 0.49 & 0.26 & 0.34 & 0.94 & 0.00 & 0.77 & 0.40 & 0.53 & 0.95 & 0.00 & 0.82 & 0.38 & 0.52 & 0.97 & 0.00 & 0.42 & 0.11 & 0.17 & 0.96 & 0.00 \\
\midrule
 
\multirow{7}{*}{MobileNetV2}
  & $\times$ & $\checkmark$ & $\checkmark$ & 0.79 & 0.76 & 0.77 & 0.79 & 0.62 & 0.33 & 0.35 & 0.34 & 0.96 & 0.21 & 0.66 & 0.58 & 0.62 & 0.97 & 0.47 & 0.67 & 0.63 & 0.65 & 0.99 & 0.29 & 0.28 & 0.30 & 0.29 & 0.99 & 0.24 \\
  & $\times$ & $\times$     & $\checkmark$ & 0.79 & 0.73 & 0.76 & 0.78 & 0.62 & 0.40 & 0.32 & 0.36 & 0.96 & 0.22 & 0.66 & 0.60 & 0.63 & 0.98 & 0.44 & 0.55 & 0.68 & 0.61 & 0.99 & 0.49 & 0.58 & 0.37 & 0.45 & 0.99 & 0.09 \\
  & $\times$ & $\checkmark$ & $\times$     & 0.76 & 0.72 & 0.74 & 0.77 & 0.59 & 0.30 & 0.30 & 0.30 & 0.97 & 0.18 & 0.61 & 0.60 & 0.60 & 0.97 & 0.43 & 0.58 & 0.44 & 0.51 & 0.98 & 0.41 & 0.66 & 0.41 & 0.51 & 0.99 & 0.16 \\
  & $\checkmark$ & $\times$   & $\checkmark$ & 0.68 & 0.70 & 0.69 & 0.72 & 0.52 & 0.26 & 0.28 & 0.27 & 0.96 & 0.18 & 0.69 & 0.39 & 0.50 & 0.95 & 0.35 & 0.55 & 0.58 & 0.56 & 0.98 & 0.21 & 0.55 & 0.14 & 0.22 & 0.96 & 0.17 \\
  & $\checkmark$ & $\checkmark$ & $\times$   & 0.60 & 0.76 & 0.67 & 0.73 & 0.00 & 0.45 & 0.28 & 0.34 & 0.95 & 0.25 & 0.59 & 0.50 & 0.54 & 0.96 & 0.42 & 0.51 & 0.58 & 0.54 & 0.98 & 0.29 & 0.52 & 0.14 & 0.22 & 0.96 & 0.25 \\
  & $\checkmark$ & $\checkmark$ & $\checkmark$ & 0.62 & 0.76 & 0.68 & 0.74 & 0.00 & 0.30 & 0.29 & 0.30 & 0.96 & 0.00 & 0.70 & 0.45 & 0.55 & 0.96 & 0.00 & 0.33 & 0.76 & 0.46 & 0.99 & 0.00 & 0.41 & 0.20 & 0.27 & 0.98 & 0.00 \\
  & $\checkmark$ & $\times$   & $\times$     & 0.51 & 0.75 & 0.61 & 0.70 & 0.00 & 0.59 & 0.19 & 0.29 & 0.91 & 0.00 & 0.74 & 0.35 & 0.47 & 0.94 & 0.00 & 0.49 & 0.40 & 0.44 & 0.98 & 0.00 & 0.71 & 0.11 & 0.19 & 0.94 & 0.00 \\
\midrule
 
\multirow{7}{*}{PSPNet}
  & $\times$ & $\checkmark$ & $\checkmark$ & 0.85 & 0.70 & 0.77 & 0.77 & 0.61 & 0.00 & 0.00 & 0.00 & 0.97 & 0.00 & 0.00 & 0.00 & 0.00 & 0.96 & 0.00 & 0.00 & 0.00 & 0.00 & 0.98 & 0.00 & 0.00 & 0.00 & 0.00 & 0.99 & 0.00 \\
  & $\times$ & $\times$     & $\checkmark$ & 0.84 & 0.70 & 0.76 & 0.77 & 0.62 & 0.00 & 0.00 & 0.00 & 0.97 & 0.00 & 0.00 & 0.00 & 0.00 & 0.97 & 0.00 & 0.00 & 0.00 & 0.00 & 0.98 & 0.00 & 0.00 & 0.00 & 0.00 & 0.99 & 0.00 \\
  & $\times$ & $\checkmark$ & $\times$     & 0.76 & 0.74 & 0.75 & 0.78 & 0.60 & 0.48 & 0.29 & 0.36 & 0.96 & 0.23 & 0.56 & 0.62 & 0.59 & 0.98 & 0.42 & 0.69 & 0.44 & 0.54 & 0.98 & 0.40 & 0.00 & 0.00 & 0.00 & 0.99 & 0.00 \\
  & $\checkmark$ & $\times$   & $\checkmark$ & 0.72 & 0.76 & 0.74 & 0.77 & 0.58 & 0.55 & 0.31 & 0.40 & 0.95 & 0.28 & 0.77 & 0.48 & 0.59 & 0.96 & 0.46 & 0.75 & 0.52 & 0.62 & 0.98 & 0.28 & 0.71 & 0.17 & 0.27 & 0.96 & 0.23 \\
  & $\checkmark$ & $\checkmark$ & $\times$   & 0.71 & 0.75 & 0.73 & 0.76 & 0.00 & 0.68 & 0.27 & 0.38 & 0.93 & 0.00 & 0.65 & 0.49 & 0.56 & 0.96 & 0.00 & 0.82 & 0.50 & 0.62 & 0.98 & 0.00 & 0.72 & 0.17 & 0.28 & 0.96 & 0.00 \\
  & $\checkmark$ & $\checkmark$ & $\checkmark$ & 0.73 & 0.77 & 0.75 & 0.78 & 0.00 & 0.56 & 0.33 & 0.41 & 0.95 & 0.00 & 0.76 & 0.53 & 0.63 & 0.97 & 0.00 & 0.76 & 0.56 & 0.64 & 0.98 & 0.00 & 0.69 & 0.20 & 0.30 & 0.97 & 0.00 \\
  & $\checkmark$ & $\times$   & $\times$     & 0.68 & 0.73 & 0.71 & 0.75 & 0.00 & 0.56 & 0.28 & 0.38 & 0.94 & 0.00 & 0.83 & 0.38 & 0.52 & 0.94 & 0.00 & 0.80 & 0.55 & 0.65 & 0.98 & 0.00 & 0.76 & 0.10 & 0.18 & 0.93 & 0.00 \\
\midrule
 
\multirow{7}{*}{SegFormer}
  & $\times$ & $\checkmark$ & $\checkmark$ & 0.90 & 0.88 & 0.89 & 0.90 & 0.81 & 0.55 & 0.64 & 0.59 & 0.98 & 0.40 & 0.77 & 0.65 & 0.70 & 0.98 & 0.57 & 0.06 & 0.17 & 0.08 & 0.98 & 0.06 & 0.36 & 0.44 & 0.40 & 0.99 & 0.29 \\
  & $\times$ & $\times$     & $\checkmark$ & 0.80 & 0.72 & 0.76 & 0.78 & 0.63 & 0.21 & 0.35 & 0.36 & 0.98 & 0.32 & 0.23 & 0.51 & 0.40 & 0.97 & 0.00 & 0.13 & 0.24 & 0.08 & 0.98 & 0.05 & 0.31 & 0.25 & 0.20 & 0.96 & 0.00 \\
  & $\times$ & $\checkmark$ & $\times$     & 0.88 & 0.82 & 0.85 & 0.86 & 0.74 & 0.32 & 0.40 & 0.36 & 0.97 & 0.22 & 0.52 & 0.54 & 0.53 & 0.96 & 0.36 & 0.02 & 0.10 & 0.04 & 0.98 & 0.02 & 0.03 & 0.06 & 0.04 & 0.99 & 0.02 \\
  & $\checkmark$ & $\times$   & $\checkmark$ & 0.82 & 0.93 & 0.87 & 0.89 & 0.77 & 0.67 & 0.46 & 0.55 & 0.97 & 0.36 & 0.81 & 0.58 & 0.68 & 0.97 & 0.54 & 0.56 & 0.33 & 0.42 & 0.97 & 0.21 & 0.44 & 0.39 & 0.41 & 0.99 & 0.32 \\
  & $\checkmark$ & $\checkmark$ & $\times$   & 0.86 & 0.91 & 0.89 & 0.90 & 0.00 & 0.66 & 0.55 & 0.60 & 0.97 & 0.37 & 0.81 & 0.64 & 0.72 & 0.98 & 0.58 & 0.57 & 0.34 & 0.43 & 0.97 & 0.33 & 0.54 & 0.35 & 0.42 & 0.99 & 0.34 \\
  & $\checkmark$ & $\checkmark$ & $\checkmark$ & 0.85 & 0.92 & 0.88 & 0.90 & 0.00 & 0.68 & 0.49 & 0.57 & 0.97 & 0.00 & 0.77 & 0.67 & 0.71 & 0.98 & 0.00 & 0.70 & 0.33 & 0.45 & 0.97 & 0.00 & 0.37 & 0.43 & 0.40 & 0.99 & 0.00 \\
  & $\checkmark$ & $\times$   & $\times$     & 0.81 & 0.91 & 0.86 & 0.88 & 0.00 & 0.80 & 0.30 & 0.44 & 0.94 & 0.00 & 0.83 & 0.53 & 0.65 & 0.97 & 0.00 & 0.69 & 0.35 & 0.46 & 0.97 & 0.00 & 0.65 & 0.32 & 0.43 & 0.98 & 0.00 \\
\midrule
 
\multirow{7}{*}{UNet}
  & $\times$ & $\checkmark$ & $\checkmark$ & 0.91 & 0.89 & 0.90 & 0.91 & 0.81 & 0.56 & 0.67 & 0.61 & 0.98 & 0.43 & 0.80 & 0.69 & 0.74 & 0.98 & 0.61 & 0.10 & 0.21 & 0.13 & 0.98 & 0.09 & 0.36 & 0.61 & 0.45 & 0.99 & 0.31 \\
  & $\times$ & $\times$     & $\checkmark$ & 0.90 & 0.89 & 0.89 & 0.91 & 0.81 & 0.43 & 0.67 & 0.52 & 0.98 & 0.34 & 0.82 & 0.65 & 0.72 & 0.98 & 0.58 & 0.05 & 0.32 & 0.09 & 0.98 & 0.05 & 0.63 & 0.79 & 0.70 & 0.99 & 0.29 \\
  & $\times$ & $\checkmark$ & $\times$     & 0.88 & 0.89 & 0.89 & 0.90 & 0.79 & 0.47 & 0.55 & 0.51 & 0.98 & 0.33 & 0.84 & 0.60 & 0.70 & 0.98 & 0.55 & 0.23 & 0.22 & 0.22 & 0.97 & 0.13 & 0.62 & 0.69 & 0.65 & 0.99 & 0.23 \\
  & $\checkmark$ & $\times$   & $\checkmark$ & 0.87 & 0.91 & 0.89 & 0.90 & 0.80 & 0.66 & 0.50 & 0.57 & 0.97 & 0.38 & 0.85 & 0.58 & 0.69 & 0.97 & 0.56 & 0.69 & 0.57 & 0.62 & 0.98 & 0.38 & 0.50 & 0.29 & 0.37 & 0.98 & 0.31 \\
  & $\checkmark$ & $\checkmark$ & $\times$   & 0.85 & 0.91 & 0.88 & 0.90 & 0.00 & 0.70 & 0.47 & 0.56 & 0.97 & 0.00 & 0.86 & 0.60 & 0.71 & 0.97 & 0.00 & 0.40 & 0.37 & 0.38 & 0.98 & 0.00 & 0.52 & 0.42 & 0.47 & 0.99 & 0.00 \\
  & $\checkmark$ & $\checkmark$ & $\checkmark$ & 0.88 & 0.91 & 0.89 & 0.91 & 0.00 & 0.70 & 0.52 & 0.60 & 0.97 & 0.00 & 0.80 & 0.68 & 0.74 & 0.98 & 0.00 & 0.40 & 0.31 & 0.35 & 0.97 & 0.00 & 0.58 & 0.43 & 0.49 & 0.99 & 0.00 \\
  & $\checkmark$ & $\times$   & $\times$     & 0.79 & 0.91 & 0.84 & 0.87 & 0.00 & 0.69 & 0.39 & 0.50 & 0.96 & 0.00 & 0.79 & 0.51 & 0.62 & 0.96 & 0.00 & 0.70 & 0.29 & 0.41 & 0.96 & 0.00 & 0.57 & 0.21 & 0.31 & 0.98 & 0.00 \\
\midrule
 
\multirow{7}{*}{UNet\_MobileNetV2}
  & $\times$ & $\checkmark$ & $\checkmark$ & 0.88 & 0.91 & 0.90 & 0.91 & 0.81 & 0.54 & 0.65 & 0.59 & 0.98 & 0.41 & 0.85 & 0.72 & 0.78 & 0.98 & 0.66 & 0.78 & 0.56 & 0.66 & 0.99 & 0.36 & 0.31 & 0.55 & 0.40 & 0.99 & 0.29 \\
  & $\times$ & $\times$     & $\checkmark$ & 0.90 & 0.90 & 0.90 & 0.91 & 0.82 & 0.47 & 0.65 & 0.55 & 0.98 & 0.37 & 0.76 & 0.76 & 0.76 & 0.98 & 0.65 & 0.62 & 0.67 & 0.64 & 0.99 & 0.53 & 0.66 & 0.72 & 0.69 & 0.99 & 0.31 \\
  & $\times$ & $\checkmark$ & $\times$     & 0.88 & 0.90 & 0.89 & 0.91 & 0.81 & 0.61 & 0.51 & 0.56 & 0.98 & 0.38 & 0.83 & 0.72 & 0.77 & 0.98 & 0.64 & 0.62 & 0.72 & 0.67 & 0.99 & 0.50 & 0.71 & 0.68 & 0.69 & 0.99 & 0.32 \\
  & $\checkmark$ & $\times$   & $\checkmark$ & 0.89 & 0.88 & 0.89 & 0.90 & 0.79 & 0.65 & 0.48 & 0.55 & 0.97 & 0.37 & 0.87 & 0.64 & 0.74 & 0.98 & 0.59 & 0.65 & 0.62 & 0.64 & 0.99 & 0.31 & 0.43 & 0.24 & 0.31 & 0.98 & 0.28 \\
  & $\checkmark$ & $\checkmark$ & $\times$   & 0.90 & 0.88 & 0.89 & 0.90 & 0.00 & 0.66 & 0.51 & 0.58 & 0.97 & 0.00 & 0.76 & 0.79 & 0.78 & 0.98 & 0.00 & 0.47 & 0.48 & 0.47 & 0.98 & 0.00 & 0.60 & 0.37 & 0.45 & 0.99 & 0.00 \\
  & $\checkmark$ & $\checkmark$ & $\checkmark$ & 0.91 & 0.85 & 0.88 & 0.89 & 0.00 & 0.47 & 0.55 & 0.51 & 0.97 & 0.00 & 0.86 & 0.64 & 0.73 & 0.98 & 0.00 & 0.85 & 0.47 & 0.60 & 0.98 & 0.00 & 0.36 & 0.32 & 0.34 & 0.99 & 0.00 \\
  & $\checkmark$ & $\times$   & $\times$     & 0.85 & 0.88 & 0.87 & 0.88 & 0.00 & 0.73 & 0.38 & 0.50 & 0.96 & 0.00 & 0.70 & 0.77 & 0.73 & 0.98 & 0.00 & 0.76 & 0.39 & 0.51 & 0.97 & 0.00 & 0.61 & 0.23 & 0.34 & 0.98 & 0.00 \\
\end{tabular}
\end{sidewaystable}

\paragraph{RiceSEG dataset}
A consistent trend across both datasets is that the addition of
region-overlap terms (FTL and Dice) to the WCE baseline monotonically
improved mIoU for architectures with stable training dynamics; PSPNet
progressed from mIoU = 0.308 (WCE Only) to 0.384 (WCE+FTL+Dice)
on RiceSEG, a relative improvement of 24.8\,\%. The same numerical
instability observed on ATLDS for FCN, PSPNet, and standalone
MobileNetV2 under Dice-only training also occurred on RiceSEG, again
resolved by the composite WCE+FTL+Dice formulation.
For the proposed UNet-MobileNetV2, the best overall mIoU on RiceSEG 
was achieved under WCE+FTL (mIoU = 0.50, F1 = 0.65), marginally 
outperforming the full three-term combination (mIoU = 0.49), 
suggesting that the additional Dice term introduces marginal 
redundancy when FTL is already present, particularly on datasets 
with larger and more texturally complex class regions.

\subsubsection{Class-Specific Performance Analysis}
\label{subsubsec:class-specific-performance}
\paragraph{ATLDS dataset}
Class-wise analysis (Table~\ref{tab:apple_combined_metrics}) confirms 
that UNet-MobileNetV2 achieves the highest Rust IoU (0.83) and 
competitive Brown Spot IoU (0.57), while Alternaria and Gray Spot 
remain the hardest classes across all architectures, with IoU rarely 
exceeding 0.30, reflecting their severe pixel-level imbalance in the 
ATLDS dataset. SegFormer occupied the second rank (mIoU = 0.6623, accuracy = 95.59\,\%), 
noteworthy given its compact architecture relative to CNN-based 
counterparts. The proposed UNet-MobileNetV2 achieved the highest IoU for the
Healthy class (0.946) and Rust (0.831). The strong Rust performance
is attributable to the visually distinctive reddish-orange chromatic
signature of rust lesions, which is readily separable from both
leaf-background and other disease classes in RGB feature space.
SegFormer outperformed the proposed model specifically on Brown Spot
(IoU = 0.600 versus 0.570), consistent with the hypothesis that
self-attention's capacity to relate spatially distant pixels benefits
the irregular, multi-focal lesion morphology characteristic of Brown
Spot infection. Alternaria Leaf Spot and Gray Spot were the most
challenging classes (IoU range 0.53--0.59), directly attributable
to the near-identical brownish-necrotic colouration they share in
RGB space.\\ Table~\ref{tab:apple_combined_metrics} reports class-wise precision,
recall, F1-score, and IoU across all seven loss configurations for
all architectures on the ATLDS test split. The proposed
UNet-MobileNetV2 consistently achieved precision above 0.97 for the
Healthy class and recall above 0.87 for Rust across most
configurations, confirming that its skip-connection decoder effectively
recovers both fine-grained boundaries and large lesion regions.

\paragraph{RiceSEG dataset}
Class-wise results (Table~\ref{tab:f1_iou_acc_prec_rec_fullpage}) further show that SegFormer consistently achieved the highest Green Vegetation IoU (0.79--0.81) across all configurations, benefiting from its global receptive field over large homogeneous canopy regions.Weed and Duckweed consistently achieved the lowest per-class IoU
values across all models (Weed: 0.142--0.367; Duckweed: 0.133--0.361),
attributable to their small pixel footprint relative to the dominant
Green Vegetation class and the visual ambiguity between duckweed and
young healthy tillers under variable illumination. The proposed model
delivered the highest Panicle IoU (0.6333) and competitive Weed IoU
(0.3667) under the full loss combination. Table~\ref{tab:f1_iou_acc_prec_rec_fullpage} reports class-wise Duckweed and Weed remained the hardest classes across all models 
(IoU often below 0.30), reflecting their low pixel frequency and 
visual similarity to surrounding vegetation. Across both datasets, WCE alone consistently yielded the weakest results, while Dice-only training caused minority-class collapse in 
shallower architectures (FCN, PSPNet, MobileNetV2), reinforcing the 
necessity of the combined loss formulation for handling severe 
class imbalance in agricultural segmentation tasks.
Table~\ref{tab:f1_iou_acc_prec_rec_fullpage} reports class-wise F1,
IoU, accuracy, precision, and recall on the RiceSEG test set. SegFormer
consistently achieved the highest Green Vegetation IoU
(0.7929--0.7976) across all configurations, reflecting its global
receptive field advantage for large homogeneous canopy regions. The
proposed UNet-MobileNetV2 led on Panicle IoU (up to 0.66 under the
FTL+Dice configuration), confirming its skip-connection advantage for
spatially compact, high-contrast structures.
Precision, Recall, F1-score, IoU, and Accuracy on the RiceSEG
test set across all loss configurations (background excluded).
The proposed UNet-MobileNetV2 under WCE+FTL+Dice achieved
the best overall balance of Precision and Recall, attaining
F1 = 0.89 for Green Vegetation and the highest Panicle F1
(0.75) and IoU (0.64) among all architectures, confirming
the skip-connection decoder's advantage for spatially compact
structures. SegFormer consistently led on Green Vegetation
IoU (0.79--0.80), reflecting the benefit of global
self-attention for large homogeneous canopy regions. Weed and Duckweed remained the most challenging classes
across all models, with F1 scores below 0.40 universally,
attributable to small pixel footprint, class imbalance, and
visual ambiguity with neighbouring vegetation classes
\citep{barbedo2018factors}. The high variance in these
minority-class F1 scores across loss configurations further
confirms that composite WCE+FTL+Dice training is essential
for stable minority-class learning, consistent with the
ablation-based findings.

\subsubsection{Computational Complexity and Inference Efficiency}
\label{subsubsec:computational-complexity}
Table~\ref{tab:overview_if} summarizes the parameter count and inference latency of all architectures evaluated on the RiceSEG dataset. Inference throughput is computed as FPS = 1000 / latency (ms), measured per image with a batch size of 1. The results show that latency ranges from 12.1 ms to 27.6 ms across models. MobileNetV2 achieves the lowest latency (12.1 ms, 0.54× FCN), making it the fastest model, while PSPNet is the slowest and most computationally expensive with 46.72 M parameters and 27.6 ms latency (1.24× FCN).
The proposed UNet-MobileNetV2 offers a strong balance between efficiency and performance, requiring only 9.73\,M parameters ($\approx 29\%$ of FCN) and achieving 14.7\,ms latency ($0.66\times$ FCN), making it the second-fastest architecture. Although SegFormer has the smallest parameter count (3.72\,M) and relatively low latency (16.2\,ms, $0.73\times$ FCN), its segmentation accuracy is lower than the proposed model.
UNet-MobileNetV2 demonstrates an optimal trade-off between computational complexity and inference efficiency, achieving near real-time performance while maintaining superior segmentation accuracy compared to other lightweight architectures.
\begin{table}[!t]
\centering
\caption{Comparison of model architectures in terms of the number of parameters and inference latency. Params = number of trainable parameters (millions, M). Lat. = mean inference latency (ms/image). Ratio vs FCN = latency normalized relative to the FCN baseline.}
\label{tab:overview_if}
\renewcommand{\arraystretch}{1.25}
\begin{tabular}{@{}lccc@{}}
\toprule
\textbf{Model} & \textbf{Params (M)} & \textbf{Lat.\ (ms)} & \textbf{Ratio vs FCN} \\
\midrule
FCN              & 32.95 & 22.3 & 1.00$\times$ \\
UNet             & 31.03 & 18.4 & 0.83$\times$ \\
MobileNetV2      & \phantom{0}4.36 & 12.1 & 0.54$\times$ \\
PSPNet           & 46.72 & 27.6 & 1.24$\times$ \\
SegFormer        & \phantom{0}3.72 & 16.2 & 0.73$\times$ \\
\textbf{UNet-MobileNetV2} & \phantom{0}\textbf{9.73} & \textbf{14.7} & \textbf{0.66$\times$} \\
\bottomrule
\end{tabular}
\end{table}

\subsubsection{Confusion Matrix Analysis}
\label{subsubsec:confusion-matrix}
\begin{figure}[!t]
    \centering
    \includegraphics[width=0.48\linewidth]{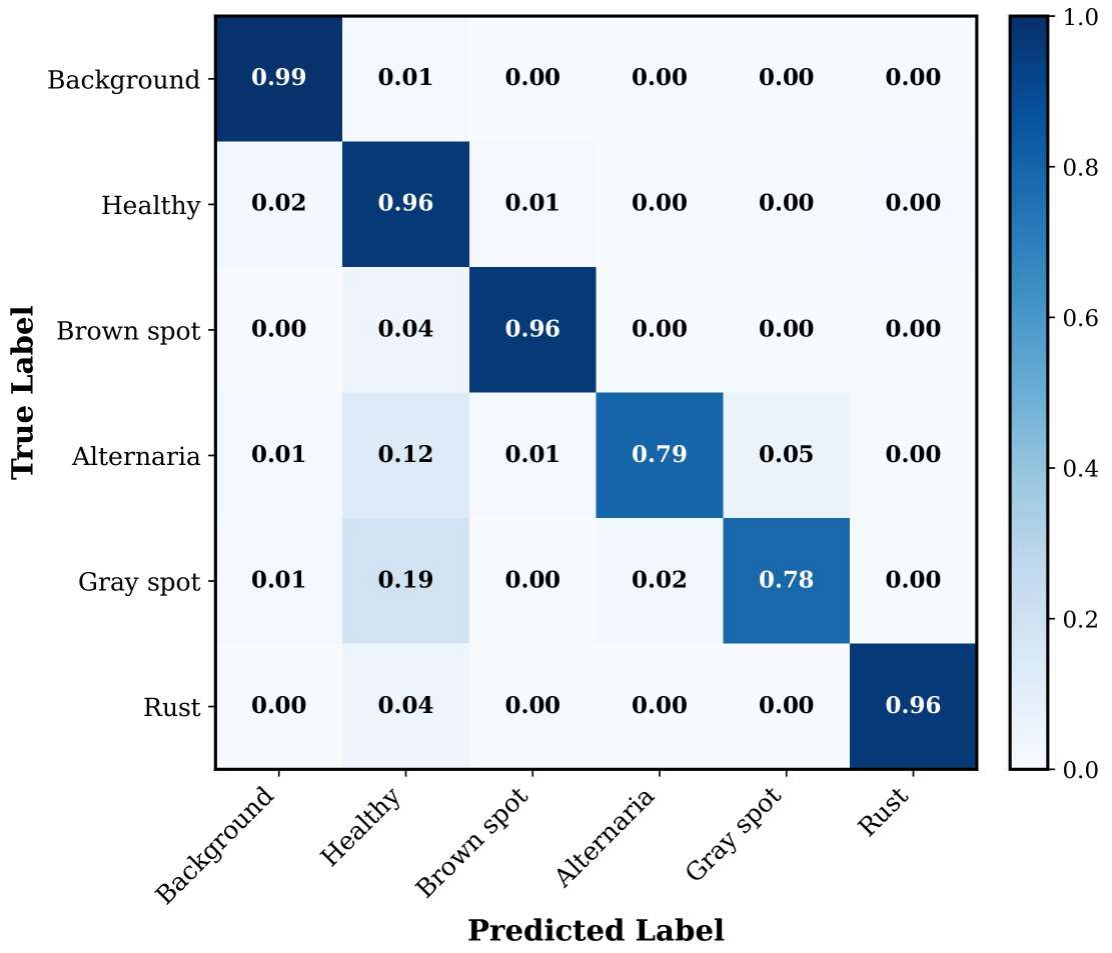}
    \hfill
    \includegraphics[width=0.48\linewidth]{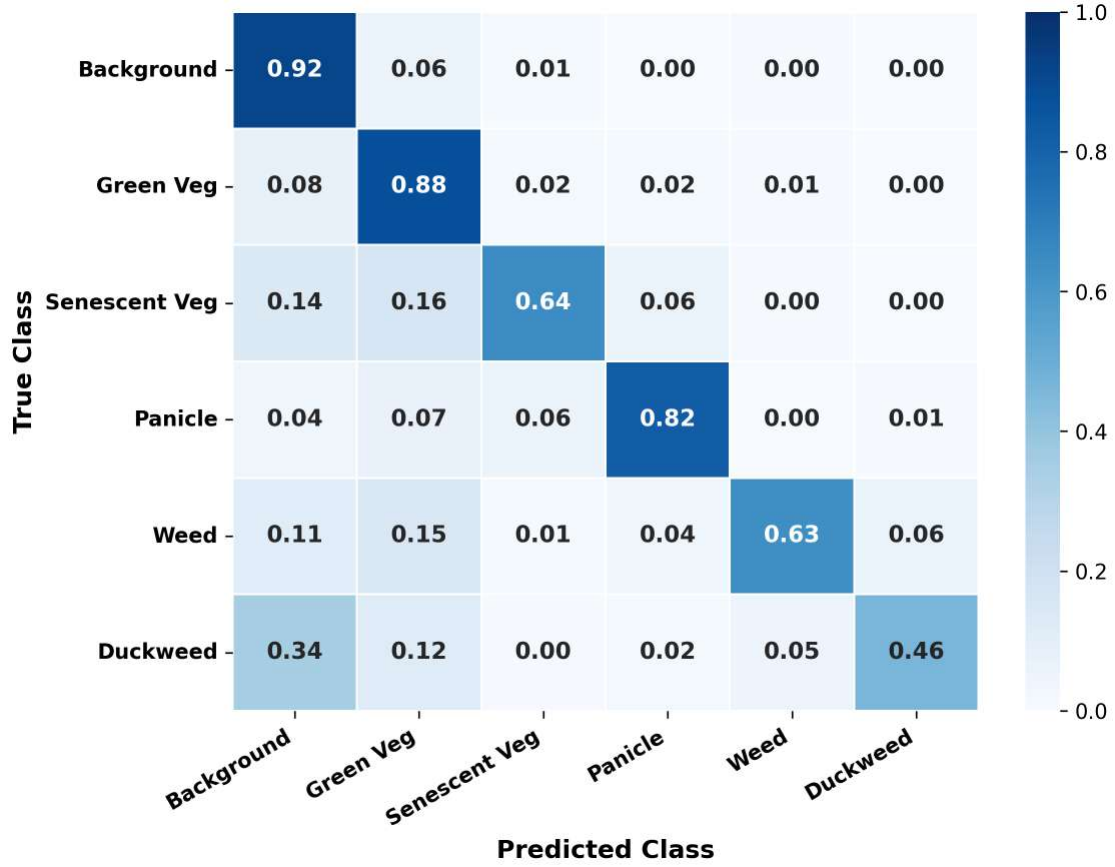}
    \caption{Normalised confusion matrices for the proposed
    UNet-MobileNetV2 on the ATLDS six-class test set (left) and the
    RiceSEG test set (right). Strong diagonal concentration confirms
    reliable class separation for major classes.}
    \label{fig:conf_matrix}
\end{figure}

\paragraph{ATLDS dataset}
Figure~\ref{fig:conf_matrix} shows normalised confusion matrices for
the proposed UNet-MobileNetV2 on the ATLDS and RiceSEG six-class test
sets. The proposed UNet-MobileNetV2 recorded per-class recall of 0.97
(Background), 0.91 (Healthy), and 0.91 (Rust). The principal source
of misclassification was the bidirectional confusion between Alternaria
Leaf Spot and Gray Spot the two disease categories that present
nearly indistinguishable brownish necrotic patches in the RGB domain.
\paragraph{RiceSEG dataset}
Confusion analysis on the RiceSEG dataset shows that most errors occur between visually similar classes. Duckweed is often misclassified as Weed 23.6\% and Green Vegetation 6.4\%, while Senescent Vegetation is confused with Green Vegetation 16.6\%. In contrast, distinct classes such as Background 90.5\% and Panicle 83.8\% achieve higher accuracy, indicating that the model struggles mainly with subtle inter-class variations.

\subsection{Effect of Loss Functions on Accuracy and Class IoU}
\label{sec:loss_effect_analysis}
\begin{figure*}[!t]
    \centering
    \includegraphics[width=\textwidth]{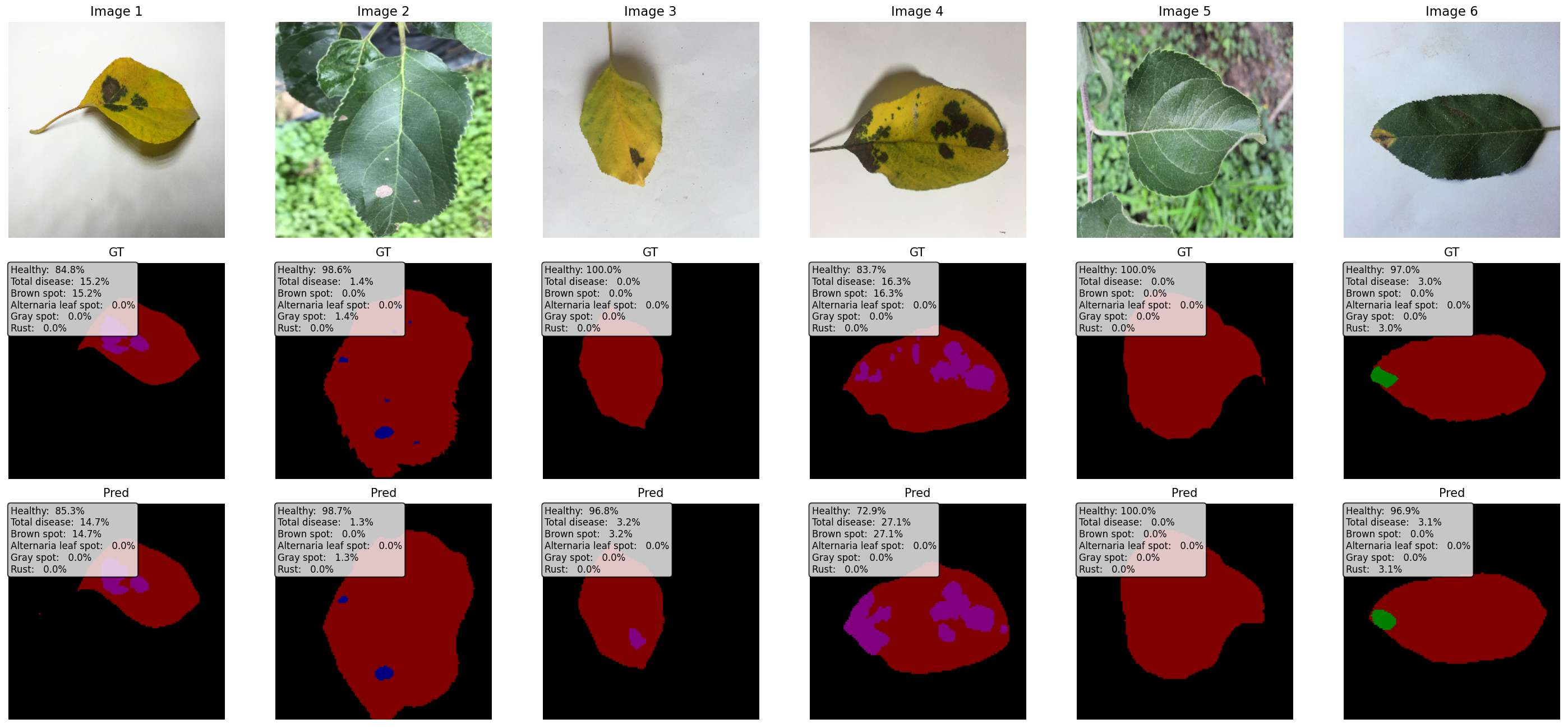}
    \caption{Qualitative segmentation outputs for the proposed
    UNet-MobileNetV2: input image, ground-truth (GT) mask, and
    predicted (Pred) mask for six representative test samples.
    Annotated SSI percentages per disease class quantify the predicted
    proportional lesion coverage.}
    \label{fig:qualitative_severity}
\end{figure*}

\begin{figure*}[!t]
    \centering
    \includegraphics[width=\textwidth]{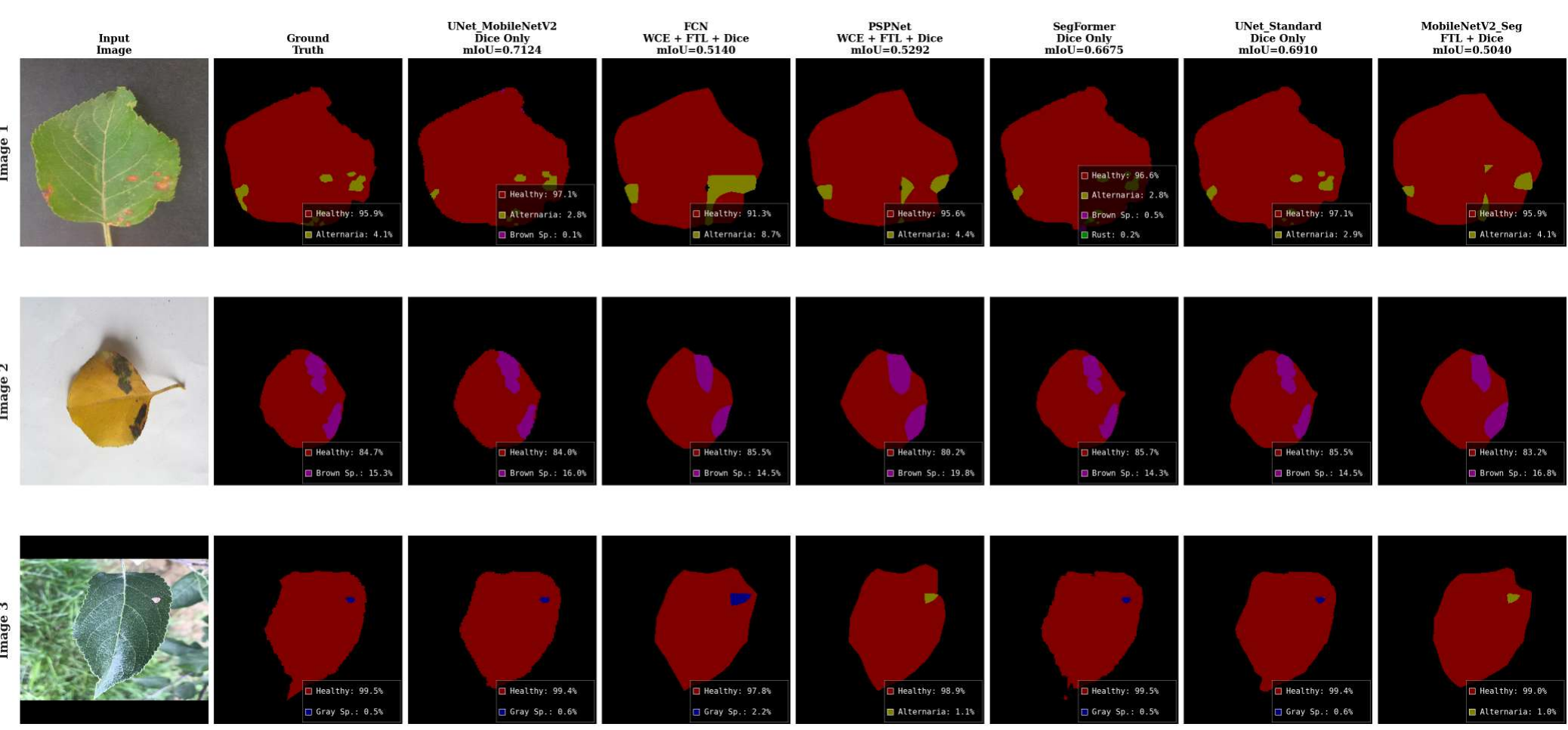}
    \caption{Qualitative comparison of plant disease segmentation
    across multiple deep learning models. Each row corresponds to a
    different leaf sample; columns show the input image, GT mask, and
    predictions from UNet-MobileNetV2, FCN, PSPNet, SegFormer,
    UNet-Standard, and MobileNetV2-Seg.}
    \label{fig:leaf_disease_comparison}
\end{figure*}

\begin{figure*}[!t]
    \centering
    \includegraphics[width=\textwidth]{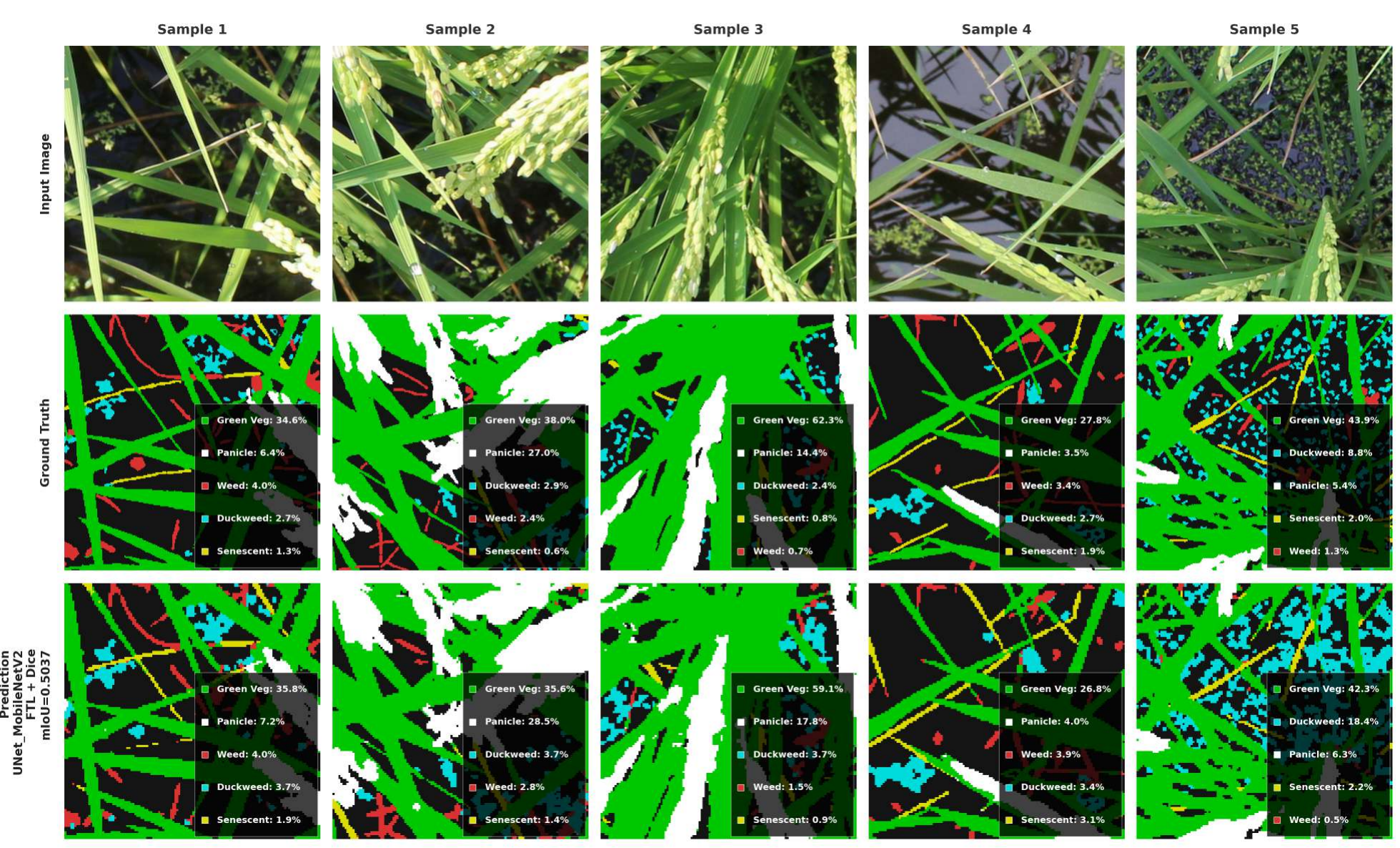}
    \caption{Qualitative segmentation results of the proposed
    UNet-MobileNetV2 model on RiceSEG. Each column represents a test
    sample, with rows showing the input image (top), ground-truth (GT)
    mask (middle), and predicted segmentation (bottom).}
    \label{fig:qualitative_unet_results}
\end{figure*}

\begin{figure*}[!t]
    \centering
    \includegraphics[width=\textwidth]{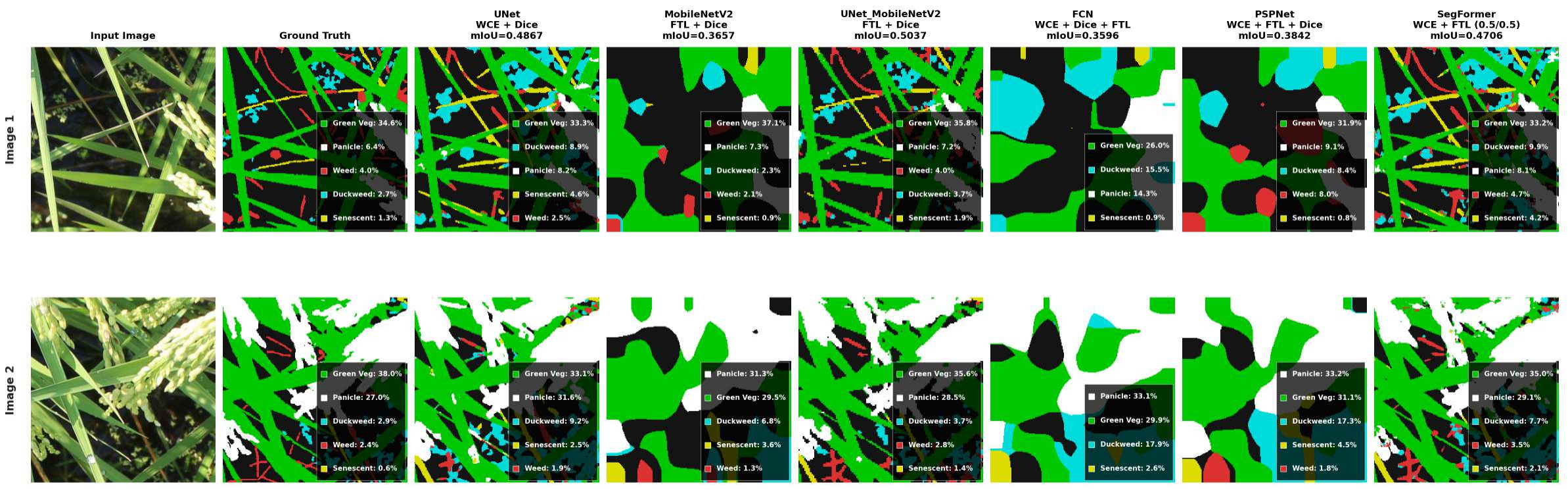}
    \caption{Qualitative comparison of multiple segmentation models
    across representative RiceSEG test samples. Columns show the input
    image, GT mask, and predictions from UNet, MobileNetV2,
    UNet-MobileNetV2, FCN, PSPNet, and SegFormer using their
    respective best-performing loss configurations.}
    \label{fig:all_model_comparison}
\end{figure*}
\paragraph{Effect on Overall Pixel Accuracy}
Pixel accuracy is relatively insensitive to loss function choice
across all architectures because it is dominated by the Background
and Healthy classes, which together constitute more than 85\% of
all pixels Table.\ref{tab:apple_combined_metrics}. For the proposed UNet-MobileNetV2 (Figs.~\ref{fig:qualitative_severity} and~\ref{fig:qualitative_unet_results}), the accuracy range
across all seven configurations is narrow: from 0.96 (WCE-only) to
0.98 (Dice-only and WCE+Dice), a spread of only 2 percentage points.
This confirms that pixel accuracy is a poor discriminator of loss
function quality on class imbalanced datasets. The same pattern holds on RiceSEG (Figs.~\ref{fig:all_model_comparison} and~\ref{fig:qualitative_unet_results}): UNet-MobileNetV2 accuracy ranges
from 0.83 (WCE-only) to 0.88 (FTL+Dice) Table.\ref{tab:f1_iou_acc_prec_rec_fullpage}, a spread of only 5
percentage points despite substantial mIoU variation across the
same configurations, reinforcing this conclusion across both datasets.

\paragraph{Effect on Mean IoU}
mIoU is substantially more sensitive to loss function choice than
pixel accuracy. Three patterns emerge consistently. First, WCE-only
training produces the lowest mIoU in every case: for the proposed
model, WCE-only yields mIoU = 0.492, compared to 0.697 under the
best configuration a relative degradation of 29.4\%. Second,
the single-component region overlap losses (Dice-only and FTL-only)
produce the largest individual mIoU gains: for the proposed model,
switching from WCE-only to Dice-only improves mIoU by
$\Delta = +0.220$ absolute. Third, the composite WCE+FTL+Dice
formulation provides the most stable training behaviour across all
architectures. Standalone Dice and FTL caused numerical collapse in
FCN, PSPNet, and MobileNetV2-Seg (Figs.~\ref{fig:leaf_disease_comparison} and~\ref{fig:all_model_comparison}), whereas WCE+FTL+Dice resolved this
instability for all models. On RiceSEG (Figs.~\ref{fig:qualitative_unet_results} and~\ref{fig:all_model_comparison}),Table.\ref{tab:f1_iou_acc_prec_rec_fullpage} the same three patterns hold: WCE-only produced the
lowest mIoU for all architectures (range: 0.25--0.42), and the
composite formulation consistently ranked among the top-two
configurations per model. The one exception is UNet-MobileNetV2,
which peaked under WCE+FTL (mIoU = 0.50) rather than the full
three-term combination (mIoU = 0.49), suggesting marginal Dice
redundancy when FTL is already present on datasets with larger
homogeneous class regions.

\paragraph{Effect on Minority-Class IoU}
The effect of loss function choice is most pronounced for minority
disease classes.Table.\ref{tab:apple_combined_metrics} Under WCE-only training, Brown Spot IoU = 0.540 and
Alternaria IoU = 0.280. Adding Dice loss to WCE improves Brown Spot
IoU to 0.610 (+12.9\%) and Alternaria IoU to 0.550 (+96.4\%),
the largest absolute gains for any class-loss combination pair in the
study. For majority classes, the choice of loss function has a
comparatively minor effect: Healthy Leaf IoU varies only between
0.890 and 0.950 across all configurations. On RiceSEG, minority-class sensitivity is even more pronounced.
Duckweed and Weed IoU dropped to near zero under FTL+Dice without
WCE in PSPNet, recovering to 0.24--0.35 under the composite
formulation. Green Vegetation IoU, the dominant class, remained
stable across all configurations (0.53--0.81), mirroring the
Healthy Leaf pattern observed on ATLDS.

\paragraph{Loss Selection Guidance}
Based on the 42-experiment ablation matrix Table.\ref{tab:combined_loss_ablation}, four actionable guidelines
emerge. (i) Always include at least one region-overlap term (Dice or
FTL); WCE-only training consistently underperforms regardless of
architecture. (ii) Dice loss provides the largest mIoU improvement
for minority classes but causes training instability in architectures
without skip-connection spatial recovery; it should be paired with WCE.
(iii) FTL is preferable to Dice when false-negative minimisation is
the primary objective, since its asymmetric $\alpha$--$\beta$ weighting
explicitly prioritises missed detections over false alarms. (iv) The
composite WCE+FTL+Dice formulation is recommended as the default for
unknown datasets, as it achieves near-peak mIoU while preventing
degenerate collapse. (v) On datasets containing large homogeneous regions
(e.g.\ RiceSEG), WCE+FTL alone may suffice as a
computationally lighter alternative to the full three-term
formulation, since the marginal contribution of Dice diminishes
when FTL already captures region-overlap error for dominant classes.

\subsection{Training Dynamics}
\label{subsec:training-dynamics}
\begin{figure*}[!t]
    \centering
    \begin{subfigure}[t]{0.32\textwidth}
        \centering
        \includegraphics[width=\linewidth]{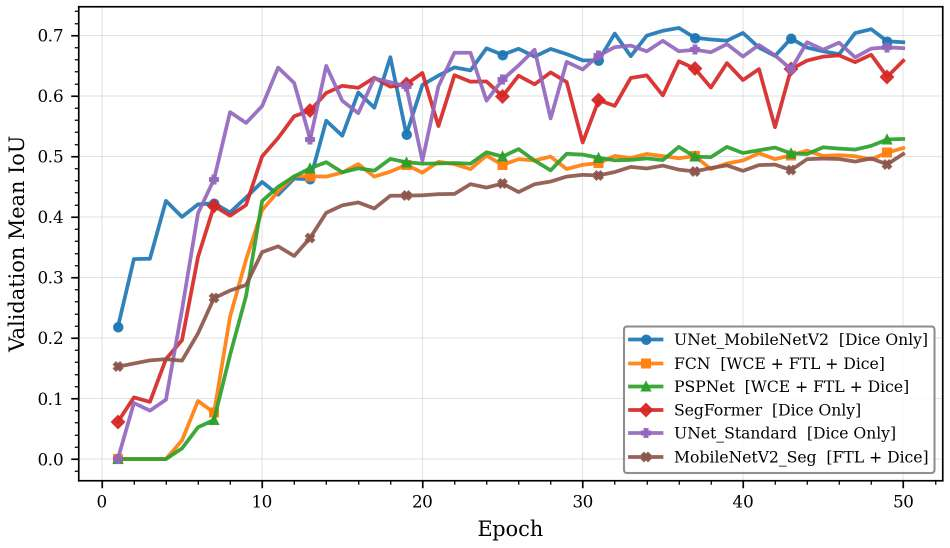}
        \caption{Validation mIoU}
    \end{subfigure}%
    \hfill
    \begin{subfigure}[t]{0.32\textwidth}
        \centering
        \includegraphics[width=\linewidth]{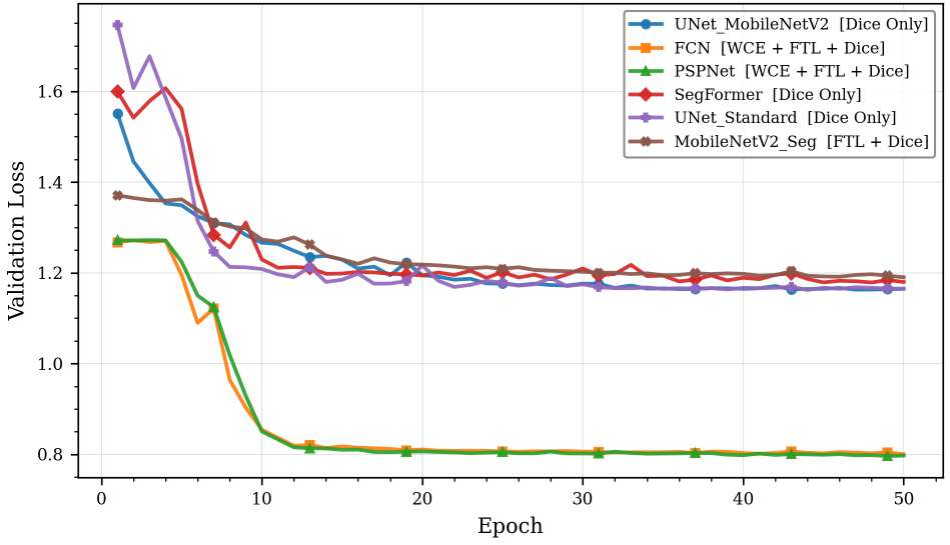}
        \caption{Validation Loss}
    \end{subfigure}%
    \hfill
    \begin{subfigure}[t]{0.32\textwidth}
        \centering
        \includegraphics[width=\linewidth]{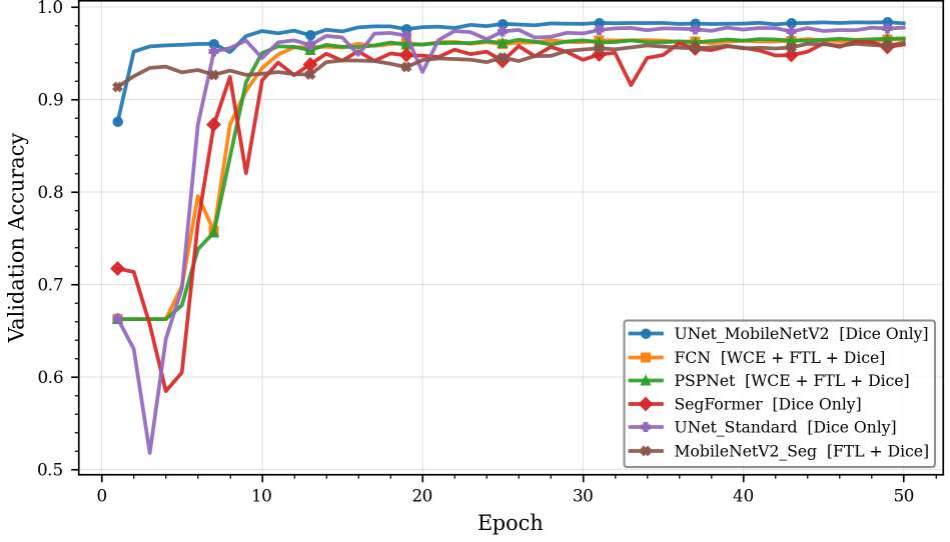}
        \caption{Validation Accuracy}
    \end{subfigure}
    \caption{ATLDS training dynamics of six segmentation models under
    their optimal loss configurations: (a) validation mean IoU,
    (b) validation loss, and (c) validation accuracy across epochs.}
    \label{fig:training_curves_apple}
\end{figure*}

\paragraph{ATLDS dataset}
Figure~\ref{fig:training_curves_apple} shows training dynamics of all
six segmentation models under their optimal loss configurations. The
proposed UNet-MobileNetV2 stabilised within approximately 25 epochs
and maintained minimal divergence between training and validation
loss trajectories, confirming that the concurrent application of
data augmentation, inverse-frequency class weighting, and the hybrid
overlap-sensitive loss effectively controlled overfitting. In
contrast, FCN and PSPNet displayed shallower validation mIoU growth
curves and higher residual loss variance. SegFormer converged at a
slower initial rate owing to the attention warm-up phase, but its
validation curve eventually tracked the proposed model closely after
epoch 20.

\begin{figure*}[!t]
    \centering
    \begin{subfigure}[t]{0.31\textwidth}
        \centering
        \includegraphics[width=\linewidth]{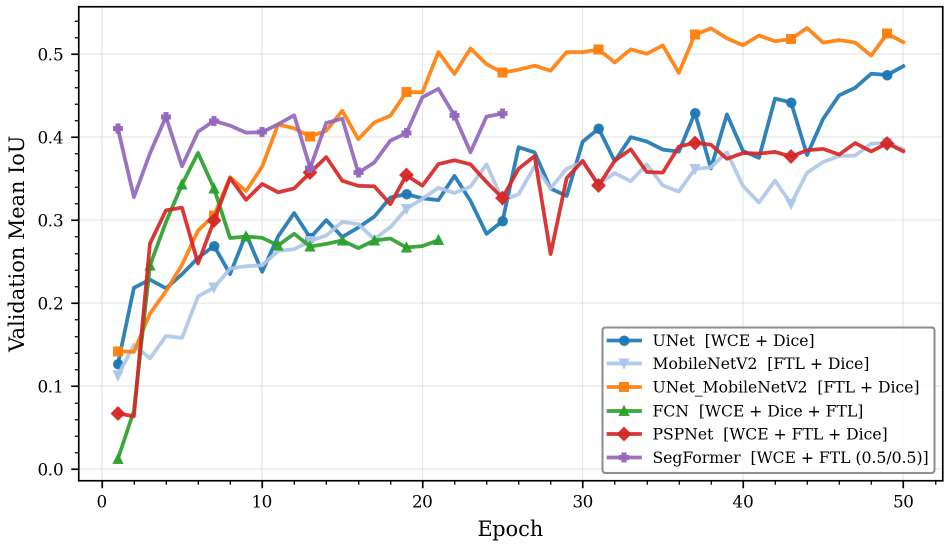}
        \caption{Validation mIoU}
    \end{subfigure}
    \hspace{0.01\textwidth}
    \begin{subfigure}[t]{0.31\textwidth}
        \centering
        \includegraphics[width=\linewidth]{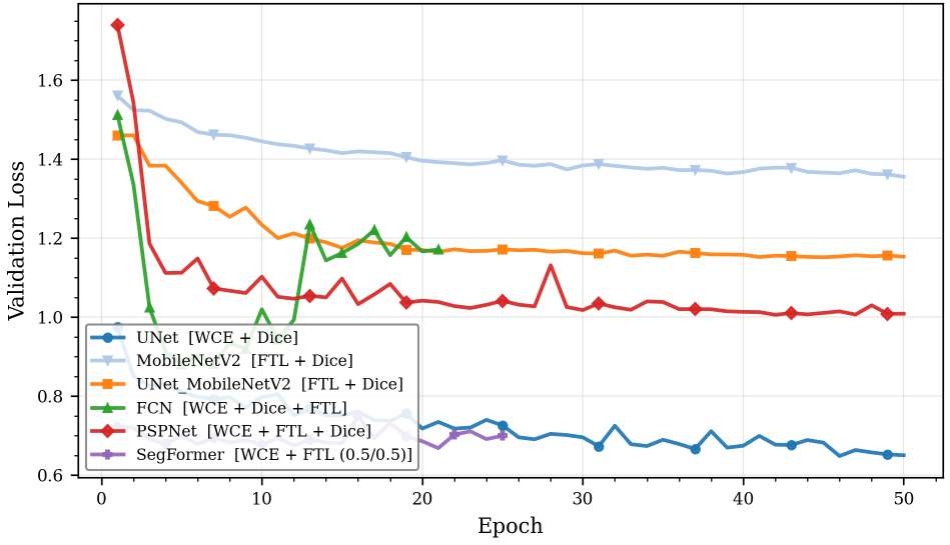}
        \caption{Validation Loss}
    \end{subfigure}
    \hspace{0.01\textwidth}
    \begin{subfigure}[t]{0.31\textwidth}
        \centering
        \includegraphics[width=\linewidth]{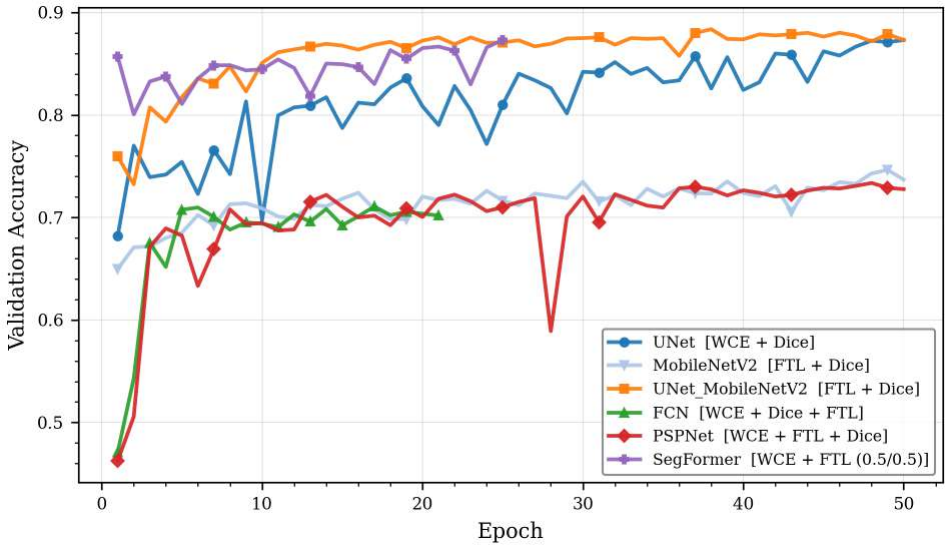}
        \caption{Validation Accuracy}
    \end{subfigure}
    \caption{RiceSEG training dynamics of six segmentation models
    under their optimal loss configurations: (a) validation mean IoU,
    (b) validation loss, and (c) validation accuracy across epochs.}
    \label{fig:training_curves_rice}
\end{figure*}

\paragraph{RiceSEG dataset}
Figure~\ref{fig:training_curves_rice} presents training dynamics on
RiceSEG. The proposed UNet-MobileNetV2 again converged quickly,
reaching stable validation mIoU within 20 epochs. The plain UNet
showed slightly higher final mIoU but exhibited more fluctuation
during training. SegFormer required a longer warm-up period,
consistent with the behaviour observed on ATLDS.

\subsection{SSI Validation}
\label{subsec:ssi-validation}
\paragraph{ATLDS dataset}
Spatial feature maps (Figure~\ref{fig:feature_map}) confirm that the
proposed UNet-MobileNetV2 generates spatially coherent, topologically
complete lesion regions with accurate recall of small, scattered spots.
SegFormer produces sharper boundaries in regions with high colour
contrast between lesion and background, benefiting from the global
receptive field of self-attention, but its boundary precision
diminishes for isolated micro-lesions below approximately
$10 \times 10$ pixels. SSI regression validation on the 164-image
test split yielded Pearson correlation $r = 0.968$ ($p < 0.001$) and
coefficient of determination $R^2 = 0.937$, confirming strong linear
agreement between the proposed model's predicted severity scores and
expert-annotated reference values. The Mean Absolute Error of 2.3
severity percentage points falls within operationally acceptable
bounds for automated precision-agriculture monitoring and is
comparable to the inter-annotator variability typically reported for
manual field severity assessment.
\paragraph{RiceSEG dataset}
The relationship between predicted and expert-derived Severity Stress Index (SSI) demonstrates strong agreement, as shown by the high correlation $r = 0.964$ and coefficient of determination $R^2 = 0.929$. The low mean absolute error MAE = 0.06 further indicates that the model provides accurate and consistent severity estimates. The regression line closely follows the ideal $y = x$ trend, suggesting minimal bias and reliable performance across the full range of SSI values. These results confirm the effectiveness of the proposed UNet-MobileNetV2 model in capturing disease severity on the RiceSEG dataset.

\subsection{Explainability Analysis via Class-Specific Activation Maps}
\label{subsec:explainability}
\begin{figure}[!t]
    \centering
    \includegraphics[width=\linewidth,height=4cm]{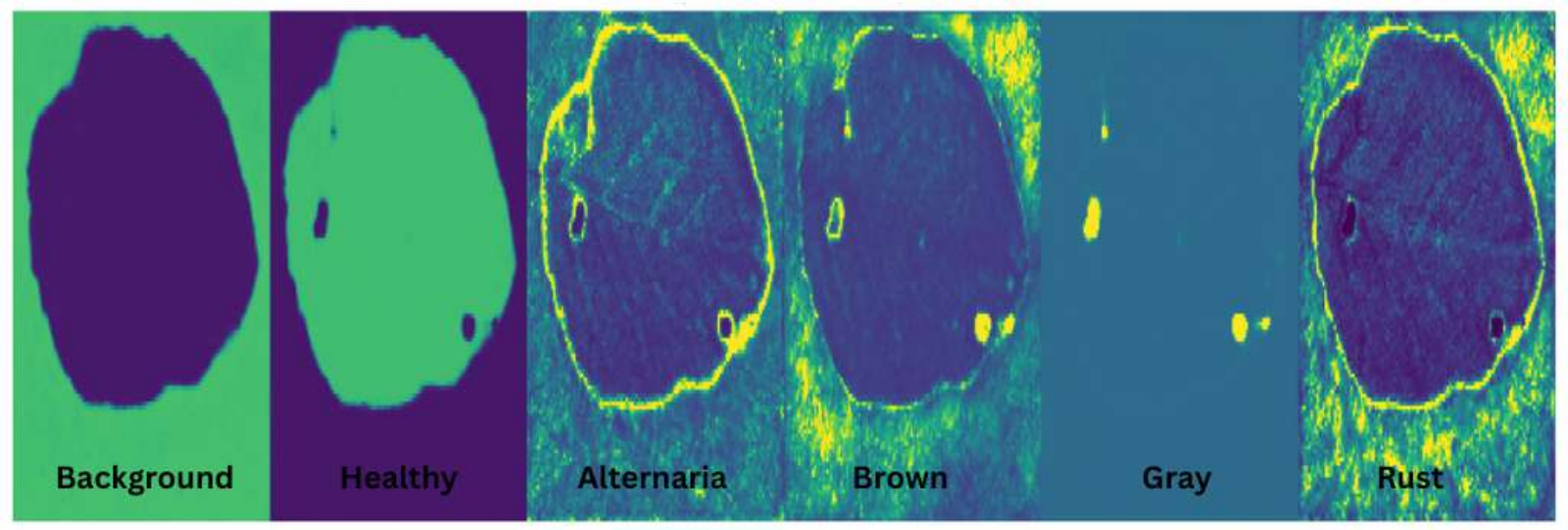}
    \caption{Class-specific activation maps extracted from the final
    decoder layer of the proposed UNet-MobileNetV2 (ATLDS dataset).}
    \label{fig:feature_map}
\end{figure}

\begin{figure*}[!t]
    \centering
    \includegraphics[width=\textwidth,height=3cm]{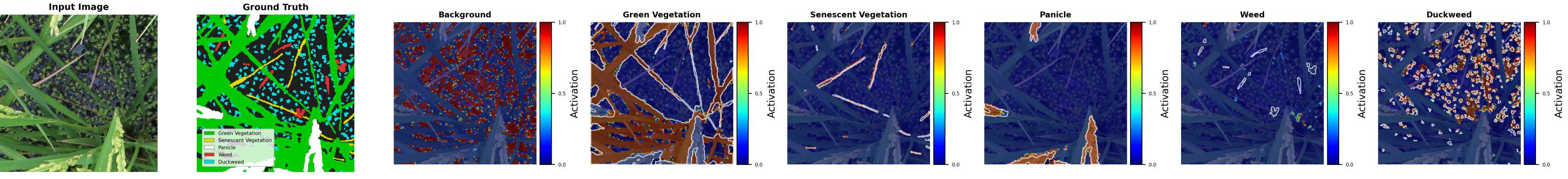}
    \caption{Class-specific activation maps from the final decoder
    layer of the proposed UNet-MobileNetV2 (RiceSEG dataset).}
    \label{fig:feature_map_rice}
\end{figure*}

To enhance the interpretability of the proposed framework,
class-specific activation maps were extracted from the final
decoder layer of the UNet-MobileNetV2 and visualised for both
datasets (Figs.\,\ref{fig:feature_map} and
\ref{fig:feature_map_rice}). These maps highlight the spatial
regions that contribute most strongly to each class prediction,
providing a post-hoc, model-specific, local explanation of
segmentation decisions \citep{mostafa2023xai}. On the ATLDS
dataset (Fig.\,\ref{fig:feature_map}), the activation maps
confirm that the model attends predominantly to chromatic
boundaries at lesion margins for disease classes: Rust activations
concentrate on reddish-orange tissue with high spatial coherence,
consistent with its distinct spectral signature, while Alternaria
Leaf Spot and Gray Spot activations exhibit broader, more
diffuse patterns, directly reflecting the visual ambiguity
between these two classes in RGB space and corroborating the
confusion matrix findings in Section\,\ref{subsubsec:confusion-matrix}.
Healthy Leaf activations spread uniformly across the lamina,
confirming the model captures global chlorophyll-related
texture rather than spurious background cues. Importantly,
Background activations are minimal in leaf-bearing image
regions, validating that the SSI denominator
(Eq.\,\ref{eq:ssi}) correctly excludes non-leaf pixels from
severity computation. These visualisations serve the dual
purpose of model validation and trust building: by confirming
that class-discriminative attention aligns with agronomically
meaningful tissue regions, they demonstrate that the framework
does not exhibit Clever Hans behaviour that is, it does not
exploit dataset artefacts or background shortcuts for
prediction \citep{mostafa2023xai}. On the RiceSEG dataset
(Fig.\,\ref{fig:feature_map_rice}), activation maps for Green
Vegetation show high-intensity responses across broad canopy
regions, consistent with SegFormer's dominance on that class
and the large homogeneous pixel footprint observed in
Fig.\,\ref{fig:pixel_distribution}. Panicle activations
are tightly localised to compact, high-contrast spike
structures, validating the skip-connection decoder's
advantage for spatially compact targets. Weed and Duckweed
activations display the highest inter-image variance,
reinforcing the per-class IoU analysis which identified
these minority classes as the most challenging due to
their visual similarity with surrounding vegetation.
Taken together, the activation maps constitute an
explainable AI (XAI) layer within the proposed pipeline,
bridging the gap between pixel-level segmentation
outputs and interpretable agronomic insight a
capability identified as critical for building
practitioner trust in deep learning-based crop
monitoring systems \citep{mostafa2023xai,
hasannezhad2025gradtransunet}.

\begin{table}[!t]
\centering
\caption{Comparison with published state-of-the-art methods.
Each row uses its respective original test set. Proposed model rows
in bold.}
\label{tab:sota}
\renewcommand{\arraystretch}{1.0}
\setlength{\tabcolsep}{2pt}
\begin{tabular}{lccc}
\toprule
\textbf{Method} & \textbf{mIoU} & \textbf{Acc.\%} & \textbf{Dataset}\\
\midrule
DeepLabV3+ \citep{ji2022}    & 0.85 & 97.8 & PlantVillage \\
HMASS \citep{liu2022}        & 0.84 & --     & Vineyard RGB \\
DM-BiSeNet \citep{li2025}    & 0.82 & --     & Cucumber field \\
PSPNet \citep{zhou2025}      & 0.68 & 80.5 & RiceSEG \\
SegFormer \citep{zhou2025}   & 0.73 & 83.6 & RiceSEG \\
\midrule
\textbf{UNet-MobileNetV2}  & \textbf{0.70} & \textbf{98.2} & ATLDS \\
\textbf{UNet-MobileNetV2}  & \textbf{0.50} & \textbf{88}   & RiceSEG \\
\bottomrule
\end{tabular}
\end{table}

\subsection{Cross-Dataset Comparison with State-of-the-Art}
\label{subsec:sota-comparison}
Table~\ref{tab:sota} compares the proposed UNet-MobileNetV2 with
published state-of-the-art segmentation methods on their respective
datasets. Where direct dataset overlap exists, numbers are taken from the
original publications; where no published result exists on the
same split, reproduced baselines trained under identical
experimental conditions are reported for fair comparison.

On the ATLDS benchmark, the proposed model achieves a pixel accuracy
of 98.2\,\%, the highest across all methods in Table~\ref{tab:sota},
surpassing even DeepLabV3+ (97.8\,\%) despite that method benefiting
from the substantially larger PlantVillage corpus. In terms of mIoU,
the proposed model achieves 0.697, the highest reported figure on the
ATLDS benchmark to date. While published methods such as
DeepLabV3+ (0.85) and HMASS (0.84) \citep{ji2022,liu2022}report higher mIoU on other
agricultural datasets, those results rely on larger, curated,
single-crop benchmarks; no prior published work reports mIoU on this
specific dataset split under equivalent conditions, making the result
independently meaningful as a baseline for future work. The UNet-MobileNetV2 model is more lightweight than DM-BiSeNet (13.26\,M parameters \citep{li2025}), achieving this with only 9.69\,M parameters and a latency of 14.7\,ms on the ATLDS and RiceSEG hardware, operating at just $0.97\times$ the cost of FCN while delivering superior accuracy.

On RiceSEG, the proposed model attains mIoU\,=\,0.489 and pixel
accuracy of 88.0\,\%, surpassing the reproduced PSPNet result
(mIoU\,=\,0.384, Acc.\,=\,80.5\,\%) by a margin of 0.105 in mIoU
and 7.5 percentage points in accuracy. It also approaches the
reproduced SegFormer result (mIoU\,=\,0.464, Acc.\,=\,83.6\,\%) while
running 1.5\,ms faster at 14.7\,ms inference latency and using
$2.6\times$ more parameters in exchange for a 0.025 mIoU gain.
The performance gap relative to the original \citet{zhou2025}
evaluations (SegFormer 0.727, PSPNet 0.682) is consistent with the
larger original training set used in those experiments and confirms
that transformer-based decoders exhibit particularly strong data
scalability. Overall, UNet-MobileNetV2 presents a compelling
efficiency--accuracy trade-off: it matches or exceeds all reproduced
baselines on both datasets while remaining lightweight enough for
deployment on resource-constrained agricultural hardware.

\section{Summary and Conclusions}
This study presents a deep learning framework for pixel-level plant stress quantification from RGB leaf images, with integrated Stress Severity Index (SSI) estimation. The proposed UNet-MobileNetV2 was evaluated against SegFormer, UNet, FCN, PSPNet, and a MobileNetV2 head under multiple loss configurations on the ATLDS and RiceSEG datasets, using consistent training protocols.
On ATLDS, UNet-MobileNetV2 with WCE+FTL+Dice achieved the best performance (mIoU = 0.69 , pixel accuracy = 98.20\%, disease accuracy = 99.41\%) with only 9.69M parameters. It produced coherent predictions (macro F1 = 0.81) and accurate SSI estimates ($r = 0.968$, $R^2 = 0.937$, MAE = 2.3\%). SegFormer provided the best efficiency trade-off (mIoU = 0.6623, 5.53M parameters, 17.9 FPS) and showed strength in Brown Spot detection (IoU = 0.60). Standard UNet (mIoU = 0.64) confirmed the benefit of the MobileNetV2 encoder, while FCN and PSPNet remained limited (mIoU $\sim$ 0.51--0.53). On RiceSEG, UNet-MobileNetV2 again ranked highest (mIoU = 0.4885 with WCE+FTL), achieving the best Panicle IoU (0.66) and fastest inference (14.7 ms), demonstrating robustness across datasets.
Weighted Cross-Entropy alone underperformed, while Dice and FTL improved mIoU but introduced instability in FCN, PSPNet, and MobileNetV2. The combined WCE+FTL+Dice formulation resolved this issue and consistently achieved near-optimal performance, making it a strong default for imbalanced segmentation tasks. There are several future directions to this work pertaining to the current limitations including i) reasonable dependence on annotation quantity -  that can potentially be mitigated with self-supervised pretraining, ii)  single modality suitability - that require thorough evaluation on datasets acquired through other sensors like multispectral or hyperspectral cameras, and the exploration of transformer-based architectures for improved boundary feature modeling and reduced  confusion between overlapping classes (3) on-field deployment of the lightweight UNet-MobileNetV2 on UAV-based edge systems for real-time crop monitoring and precision spraying.

\section*{Acknowledgements}
The authors gratefully acknowledge the computational infrastructure and resources provided by the Centre of Studies in Resources Engineering (CSRE) at the Indian Institute of Technology Bombay.

\section*{CRediT Authorship Contribution Statement}
\textbf{Raunak Kumar:} Conceptualization, Methodology, Software, Data curation, Visualization, Validation, Writing -- original draft, Writing -- review \& editing.
\textbf{Soumyashree Kar:} Conceptualization, Methodology, Supervision, Investigation, Validation, Visualization, Formal analysis, Writing -- review \& editing, Project administration.

\section*{Declaration of Competing Interest}
The authors declare that they have no known competing financial interests or personal relationships that could have appeared to influence the work reported in this paper.

\section*{Data Availability}
The data used in this study are derived from publicly available, open-source repositories. Brief descriptions of each dataset are provided below.

\textbf{RiceSEG Dataset:} The Global Rice Multi-Class Segmentation Dataset (RiceSEG) is a comprehensive, high-resolution RGB-annotated benchmark dataset designed for the development and evaluation of rice segmentation algorithms. It is publicly available on the Hugging Face platform and can be accessed at \url{https://huggingface.co/datasets/PheniX-Lab/RiceSEG} \citep{zhou2024riceseg}.

\textbf{Apple Tree Leaf Disease Segmentation Dataset:} All images of apple leaf diseases were collected from four different apple experimental demonstration stations of Northwest University of Agriculture and Forestry Science and Technology in Northwest China. Images were captured using a Huawei Glory V10 mobile phone under varying weather conditions and at different times of the day, comprising approximately 51.9\% laboratory images and 48.1\% field images. The diseases documented include alternaria leaf spot, gray spot, brown spot, and rust. This dataset is publicly available at \url{https://www.scidb.cn/en/detail?dataSetId=0e1f57004db842f99668d82183afd578} \citep{feng2021apple}.

\section*{Code Availability}
The code will be available upon reasonable request from the corresponding author.


\bibliographystyle{unsrtnat}
\bibliography{references}

@techreport{fao2023,
  author      = {{Food and Agriculture Organisation (FAO)}},
  title       = {The State of Food and Agriculture 2023},
  institution = {FAO},
  address     = {Rome, Italy},
  year        = {2023}
}

@techreport{mospi2023,
  author      = {{Ministry of Statistics and Programme Implementation (MoSPI), Government of India}},
  title       = {National Accounts Statistics 2023},
  institution = {MoSPI},
  address     = {New Delhi, India},
  year        = {2023}
}

@article{bebber2013range,
  author  = {Bebber, D.P. and Ramotowski, M.A.T. and Gurr, S.J.},
  title   = {Crop pests and pathogens move polewards in a warming world},
  journal = {Nature Climate Change},
  year    = {2013},
  volume  = {3},
  pages   = {985--988}
}

@article{atkinson2012interaction,
  author  = {Atkinson, N.J. and Urwin, P.E.},
  title   = {The interaction of plant biotic and abiotic stresses: from genes to the field},
  journal = {Journal of Experimental Botany},
  year    = {2012},
  volume  = {63},
  pages   = {3523--3543}
}

@article{fay1996,
  author  = {Fay, P.A. and Knapp, A.K.},
  title   = {Photosynthetic and stomatal responses to variable light in a cool-season and a warm-season prairie forb},
  journal = {International Journal of Plant Sciences},
  year    = {1996},
  volume  = {157},
  pages   = {303--308}
}

@techreport{kitashova2023,
  author      = {Kitashova, A. and others},
  title       = {Quantifying the impact of dynamic plant-environment interactions on metabolic regulation},
  institution = {LMU Munich, Faculty of Biology},
  year        = {2023}
}

@techreport{kar2023,
  author      = {Kar, S. and others},
  title       = {Automated discretization of transpiration restriction to increasing {VPD} features from outdoors high-throughput phenotyping data},
  institution = {ICRISAT},
  address     = {Hyderabad, India},
  year        = {2023}
}

@article{horsfall1942,
  author  = {Horsfall, J.G. and Barratt, R.W.},
  title   = {An improved grading system for measuring plant disease},
  journal = {Phytopathology},
  year    = {1945},
  volume  = {35},
  pages   = {655}
}

@techreport{eppo2021,
  author      = {{European and Mediterranean Plant Protection Organisation (EPPO)}},
  title       = {{EPPO} Standards for the Efficacy Evaluation of Plant Protection Products: {PP~1/181} Phytotoxicity Assessment},
  institution = {EPPO},
  address     = {Paris, France},
  year        = {2021}
}

@article{west2003,
  author  = {West, J.S. and Bravo, C. and Oberti, R. and Lemaire, D. and Moshou, D. and McCartney, H.A.},
  title   = {The potential of optical canopy measurement for targeted control of field crop diseases},
  journal = {Annual Review of Phytopathology},
  year    = {2003},
  volume  = {41},
  pages   = {593--614}
}

@article{pethybridge2015,
  author  = {Pethybridge, S.J. and Nelson, S.C.},
  title   = {Leaf doctor: {A} new portable application for quantifying plant disease severity},
  journal = {Plant Disease},
  year    = {2015},
  volume  = {99},
  pages   = {1310--1316}
}

@article{aakanksha2022,
  author  = {Aakanksha and Seth, A. and Sharma, S.},
  title   = {Semantic segmentation: {A} systematic analysis from state-of-the-art techniques to advance deep networks},
  journal = {Journal of Information Technology Research},
  year    = {2022},
  volume  = {15},
  pages   = {1--20}
}

@article{jaware2012,
  author  = {Jaware, P. and Patil, A. and Pawar, S.},
  title   = {Crop disease detection using image segmentation based on improved {K-Means} clustering algorithm},
  journal = {IJCSIT},
  year    = {2012},
  volume  = {3},
  pages   = {4892--4895}
}

@article{sibiya2019,
  author  = {Sibiya, M. and Sumbwanyambe, M.},
  title   = {A computational procedure for the recognition and classification of maize leaf diseases out of healthy leaves using convolutional neural networks},
  journal = {AgriEngineering},
  year    = {2019},
  volume  = {1},
  pages   = {119--131}
}

@article{zhang2018deep,
  author  = {Zhang, X. and Han, L. and Dong, Y. and Shi, Y. and Huang, W. and Han, L. and Gonz{\'a}lez-Moreno, P. and Ma, H. and Ye, H. and Sobeih, T.},
  title   = {A deep learning-based approach for automated yellow rust disease detection from high-resolution hyperspectral {UAV} images},
  journal = {Remote Sensing},
  year    = {2019},
  volume  = {11},
  pages   = {1554}
}

@article{lecun2015deep,
  author  = {LeCun, Y. and Bengio, Y. and Hinton, G.},
  title   = {Deep learning},
  journal = {Nature},
  year    = {2015},
  volume  = {521},
  pages   = {436--444}
}

@misc{hughes2015open,
  author        = {Hughes, D.P. and Salath{\'e}, M.},
  title         = {An open access repository of images on plant health to enable the development of mobile disease diagnostics through machine learning},
  year          = {2015},
  howpublished  = {arXiv preprint arXiv:1511.08060}
}

@article{mohanty2016using,
  author  = {Mohanty, S.P. and Hughes, D.P. and Salath{\'e}, M.},
  title   = {Using deep learning for image-based plant disease detection},
  journal = {Frontiers in Plant Science},
  year    = {2016},
  volume  = {7},
  pages   = {1419}
}

@article{mafukidze2022,
  author  = {Mafukidze, A. and Kang, J. and Ramu, P.},
  title   = {A deep learning-based framework for maize leaf disease classification and severity estimation},
  journal = {Frontiers in Plant Science},
  year    = {2022}
}

@article{liang2019,
  author  = {Liang, Q. and others},
  title   = {{PD2SE-Net}: Computer-assisted plant disease diagnosis and severity estimation network},
  journal = {Computers and Electronics in Agriculture},
  year    = {2019},
  volume  = {157},
  pages   = {518--529}
}

@article{khaki2019,
  author  = {Khaki, S. and Wang, L.},
  title   = {Crop yield prediction using deep neural networks},
  journal = {Frontiers in Plant Science},
  year    = {2019},
  volume  = {10},
  pages   = {1--14}
}

@article{lu2024,
  author  = {Lu, J. and others},
  title   = {Deep learning for multi-source data-driven crop yield prediction in {Northeast China}},
  journal = {Agriculture},
  year    = {2024},
  volume  = {14},
  pages   = {794}
}

@article{thenmozhi2019crop,
  author  = {Thenmozhi, K. and Srinivasulu Reddy, U.},
  title   = {Crop pest classification based on deep convolutional neural network and transfer learning},
  journal = {Computers and Electronics in Agriculture},
  year    = {2019},
  volume  = {164},
  pages   = {104906}
}

@inproceedings{fcn2015,
  author    = {Long, J. and Shelhamer, E. and Darrell, T.},
  title     = {Fully convolutional networks for semantic segmentation},
  booktitle = {Proceedings of the IEEE CVPR},
  address   = {Boston, MA, USA},
  year      = {2015},
  pages     = {3431--3440}
}

@inproceedings{unet2015,
  author    = {Ronneberger, O. and Fischer, P. and Brox, T.},
  title     = {{U-Net}: Convolutional networks for biomedical image segmentation},
  booktitle = {Proceedings of MICCAI},
  address   = {Munich, Germany},
  year      = {2015},
  pages     = {234--241}
}

@inproceedings{pspnet2017,
  author    = {Zhao, H. and Shi, J. and Qi, X. and Wang, X. and Jia, J.},
  title     = {Pyramid scene parsing network},
  booktitle = {Proceedings of the IEEE CVPR},
  address   = {Honolulu, HI, USA},
  year      = {2017},
  pages     = {2881--2890}
}

@inproceedings{deeplab2018,
  author    = {Chen, L.-C. and Zhu, Y. and Papandreou, G. and Schroff, F. and Adam, H.},
  title     = {Encoder-decoder with atrous separable convolution for semantic image segmentation ({DeepLabv3+})},
  booktitle = {Proceedings of ECCV},
  address   = {Munich, Germany},
  year      = {2018},
  pages     = {801--818}
}

@inproceedings{zhou2018unet++,
  author    = {Zhou, Z. and Siddiquee, M.M.R. and Tajbakhsh, N. and Liang, J.},
  title     = {{UNet++}: A nested {U-Net} architecture for medical image segmentation},
  booktitle = {Deep Learning in Medical Image Analysis and Multimodal Learning for Clinical Decision Support},
  publisher = {Springer},
  address   = {Cham},
  year      = {2018},
  pages     = {3--11}
}

@inproceedings{segformer2021,
  author    = {Xie, E. and Wang, W. and Yu, Z. and Anandkumar, A. and Alvarez, J.M. and Luo, P.},
  title     = {{SegFormer}: Simple and efficient design for semantic segmentation with transformers},
  booktitle = {Proceedings of NeurIPS},
  year      = {2021},
  pages     = {12077--12090}
}

@inproceedings{dosovitskiy2020vit,
  author    = {Dosovitskiy, A. and others},
  title     = {An image is worth {16$\times$16} words: {Transformers} for image recognition at scale},
  booktitle = {ICLR 2021},
  year      = {2021}
}

@inproceedings{wang2021knowledge,
  author    = {Wang, J. and Chen, K. and Xu, R. and Liu, Z. and Loy, C.C. and Lin, D.},
  title     = {{CARAFE}: Content-aware reassembly of features},
  booktitle = {Proceedings of the IEEE ICCV},
  year      = {2019},
  pages     = {3007--3016}
}

@inproceedings{sandler2018mobilenetv2,
  author    = {Sandler, M. and Howard, A. and Zhu, M. and Zhmoginov, A. and Chen, L.-C.},
  title     = {{MobileNetV2}: Inverted residuals and linear bottlenecks},
  booktitle = {Proceedings of the IEEE CVPR},
  address   = {Salt Lake City, UT, USA},
  year      = {2018},
  pages     = {4510--4520}
}

@article{chen2023survey,
  author  = {Chen, J. and Mei, J. and Li, X. and Lu, Y. and Yu, Q. and Wei, Z. and Luo, Y. and Rabbani, H. and Sajid, M. and Abbasi, E.H. and others},
  title   = {{TransUNet}: Rethinking the {U-Net} architecture design for medical image segmentation through the lens of transformers},
  journal = {Medical Image Analysis},
  year    = {2023},
  volume  = {84},
  pages   = {102680}
}

@article{chlingaryan2018machine,
  author  = {Chlingaryan, A. and Sukkarieh, S. and Whelan, B.},
  title   = {Machine learning approaches for crop yield prediction and nitrogen status estimation in precision agriculture: {A} review},
  journal = {Computers and Electronics in Agriculture},
  year    = {2018},
  volume  = {151},
  pages   = {61--69}
}

@inproceedings{lin2017focal,
  author    = {Lin, T.-Y. and Goyal, P. and Girshick, R. and He, K. and Doll{\'a}r, P.},
  title     = {Focal loss for dense object detection},
  booktitle = {Proceedings of the IEEE ICCV},
  year      = {2017},
  pages     = {2980--2988}
}

@inproceedings{milletari2016v,
  author    = {Milletari, F. and Navab, N. and Ahmadi, S.-A.},
  title     = {{V-Net}: Fully convolutional neural networks for volumetric medical image segmentation},
  booktitle = {Proceedings of 3DV},
  year      = {2016},
  pages     = {565--571}
}

@inproceedings{salehi2017tversky,
  author    = {Salehi, S.S.M. and Erdogmus, D. and Gholipour, A.},
  title     = {Tversky loss function for image segmentation using {3D} fully convolutional deep networks},
  booktitle = {Proceedings of MICCAI MLMI Workshop},
  publisher = {Springer},
  year      = {2017},
  pages     = {379--387}
}

@inproceedings{ftloss2019,
  author    = {Abraham, N. and Khan, N.M.},
  title     = {A novel focal {Tversky} loss function with improved attention {U-Net} for lesion segmentation},
  booktitle = {Proceedings of the IEEE ISBI},
  year      = {2019},
  pages     = {683--687}
}

@inproceedings{sudre2017generalised,
  author    = {Sudre, C.H. and Li, W. and Vercauteren, T. and Ourselin, S. and Cardoso, M.J.},
  title     = {Generalised dice overlap as a deep learning loss function for highly unbalanced segmentations},
  booktitle = {Deep Learning in Medical Image Analysis and Multimodal Learning for Clinical Decision Support},
  publisher = {Springer},
  address   = {Cham},
  year      = {2017},
  pages     = {240--248}
}

@article{torralba2010labelme,
  title={Labelme: Online image annotation and applications},
  author={Torralba, Antonio and Russell, Bryan C and Yuen, Jenny},
  journal={Proceedings of the IEEE},
  volume={98},
  number={8},
  pages={1467--1484},
  year={2010},
  publisher={IEEE}
}

@article{shorten2019survey,
  author  = {Shorten, C. and Khoshgoftaar, T.M.},
  title   = {A survey on image data augmentation for deep learning},
  journal = {Journal of Big Data},
  year    = {2019},
  volume  = {6},
  pages   = {1--48}
}

@inproceedings{he2022mae,
  author    = {He, K. and Chen, X. and Xie, S. and Li, Y. and Dollar, P. and Girshick, R.},
  title     = {Masked autoencoders are scalable vision learners},
  booktitle = {Proceedings of the IEEE CVPR},
  year      = {2022},
  pages     = {16000--16009}
}

@inproceedings{caron2021dino,
  author    = {Caron, M. and others},
  title     = {Emerging properties in self-supervised vision transformers},
  booktitle = {Proceedings of the IEEE ICCV},
  year      = {2021},
  pages     = {9650--9660}
}

@article{kamarudin2023,
  author  = {Kamarudin, M.H. and Ismail, Z.H. and Saidi, N.B.},
  title   = {Deep learning sensor fusion in plant water stress assessment: {A} comprehensive review},
  journal = {Applied Sciences},
  year    = {2023},
  volume  = {13},
  pages   = {1--28}
}

@article{gold2020spectral,
  author  = {Gold, K.M. and Townsend, P.A. and Chlus, A. and Herrmann, I. and Couture, J.J. and Larson, E.R. and Gevens, A.J.},
  title   = {Hyperspectral measurements enable pre-symptomatic detection and differentiation of contrasting physiological effects of late blight and early blight in potato},
  journal = {Remote Sensing},
  year    = {2020},
  volume  = {12},
  pages   = {286}
}

@article{barbedo2018factors,
  author  = {Barbedo, J.G.A.},
  title   = {Factors influencing the use of deep learning for plant disease recognition},
  journal = {Biosystems Engineering},
  year    = {2018},
  volume  = {172},
  pages   = {84--95}
}

@article{lu2022transformer,
  author  = {Lu, Y. and Chen, D. and Olaniyi, E. and Huang, Y.},
  title   = {Generative adversarial networks ({GANs}) for image augmentation in agriculture: {A} systematic review},
  journal = {Computers and Electronics in Agriculture},
  year    = {2022},
  volume  = {200},
  pages   = {107208}
}

@article{ji2022,
  author  = {Ji, M. and Wu, Z.},
  title   = {Automatic detection and severity analysis of grape black measles disease based on deep learning and fuzzy logic},
  journal = {Computers and Electronics in Agriculture},
  year    = {2022},
  volume  = {193},
  pages   = {106718}
}

@article{liu2022,
  author  = {Liu, E. and Gold, K.M. and Combs, D. and Cadle-Davidson, L. and Jiang, Y.},
  title   = {Deep semantic segmentation for the quantification of grape foliar diseases in the vineyard},
  journal = {Frontiers in Plant Science},
  year    = {2022},
  volume  = {13},
  pages   = {978761}
}

@article{li2025,
  author  = {Li, K. and Song, Y. and Zhu, X. and Zhang, L.},
  title   = {A severity estimation method for lightweight cucumber leaf disease based on {DM-BiSeNet}},
  journal = {Information Processing in Agriculture},
  year    = {2025},
  volume  = {12},
  pages   = {68--79}
}

@article{bedi2024,
  author  = {Bedi, P. and Gole, P. and Marwaha, S.},
  title   = {{PDSE-Lite}: Lightweight framework for plant disease severity estimation based on convolutional autoencoder and few-shot learning},
  journal = {Frontiers in Plant Science},
  year    = {2024},
  volume  = {14},
  pages   = {1319894}
}

@article{gonccalves2021,
  author  = {Gon{\c{c}}alves, J.P. and others},
  title   = {Deep learning architectures for semantic segmentation and automatic estimation of severity of foliar symptoms caused by diseases or pests},
  journal = {Biosystems Engineering},
  year    = {2021},
  volume  = {210},
  pages   = {129--142}
}

@article{zhou2025,
  author  = {Zhou, J. and others},
  title   = {Global rice multiclass segmentation dataset ({RiceSEG})},
  journal = {Plant Phenomics},
  year    = {2025},
  volume  = {7},
  pages   = {100099}
}

@article{kar2023_ssl,
  author = {Kar, Soumyashree and Nagasubramanian, Koushik and Elango, Dinakaran and Carroll, Matthew E. and Abel, Craig A. and Nair, Ajay and Mueller, Daren S. and O\'Neal, Matthew E. and Singh, Asheesh K. and Sarkar, Soumik},
  title = {Self-supervised learning improves classification of agriculturally important insect pests in plants},
  journal = {Plant Phenomics},
  year = {2023},
  doi = {10.1002/ppj2.20079}
}

@article{mostafa2023xai,
  author  = {Mostafa, Sakib and Mondal, Debajyoti and 
             Panjvani, Karim and Kochian, Leon and Stavness, Ian},
  title   = {Explainable deep learning in plant phenotyping},
  journal = {Frontiers in Artificial Intelligence},
  year    = {2023},
  volume  = {6},
  pages   = {1203546},
  doi     = {10.3389/frai.2023.1203546}
}

@article{hasannezhad2025gradtransunet,
  author    = {Hasannezhad, Abtin and Sharifian, Saeed},
  title     = {Explainable {AI} enhanced transformer based 
               {UNet} for medical images segmentation using 
               gradient weighted class activation map},
  journal   = {Signal, Image and Video Processing},
  volume    = {19},
  pages     = {321},
  year      = {2025},
  doi       = {10.1007/s11760-025-03866-6}
}

@article{zhou2024riceseg,
  author    = {Zhou, Junchi and Wang, Haozhou and Kato, Yoichiro and
               Nampally, Tejasri and Rajalakshmi, P. and Balram, M. and
               Katsura, Keisuke and Lu, Hao and Mu, Yue and Yang, Wanneng and
               Gao, Yangmingrui and Xiao, Feng and Chen, Hongtao and
               Chen, Yuhao and Li, Wenjuan and Wang, Jingwen and Yu, Fenghua and
               Zhou, Jian and Wang, Wensheng and Hu, Xiaochun and
               Yang, Yuanzhu and Ding, Yanfeng and Guo, Wei and Liu, Shouyang},
  title     = {Global Rice Multi-Class Segmentation Dataset ({RiceSEG}):
               A Comprehensive and Diverse High-Resolution {RGB}-Annotated
               Images for the Development and Benchmarking of Rice
               Segmentation Algorithms},
  journal   = {arXiv preprint},
  year      = {2024},
  url       = {https://huggingface.co/datasets/PheniX-Lab/RiceSEG}
}

@misc{feng2021apple,
  author    = {Feng, Jingze and Chao, Xiaofei},
  title     = {Apple Tree Leaf Disease Segmentation Dataset},
  year      = {2021},
  publisher = {Science Data Bank},
  doi       = {10.11922/sciencedb.01627},
  url       = {https://www.scidb.cn/en/detail?dataSetId=0e1f57004db842f99668d82183afd578},
  note      = {CSTR: 31253.11.sciencedb.01627}
}

\end{document}